\documentclass[sigconf]{acmart}
\usepackage{graphicx} 
\usepackage{amsmath} 
\usepackage{booktabs} 
\usepackage{tabularx} 
\usepackage{multirow} 
\usepackage{subcaption} 
\usepackage{caption} 
\usepackage{xpatch} 
\usepackage{pgfplots} 
\pgfplotsset{compat=1.16}
\usepackage{tikz} 

\AtBeginDocument{%
  }

\setcopyright{acmlicensed}
\copyrightyear{2018}
\acmYear{2018}
\acmDOI{XXXXXXX.XXXXXXX}
\acmConference[Conference acronym 'XX]{Make sure to enter the correct
  conference title from your rights confirmation email}{June 03--05,
  2018}{Woodstock, NY}
\acmISBN{978-1-4503-XXXX-X/2018/06}

\begin{document}

\title{First RAG, Second SEG: A Training-Free Paradigm for Camouflaged Object Detection}

\author{Wutao Liu}
\affiliation{%
	\institution{Nanjing University of Aeronautics and Astronautics}
	\city{Nanjing}
	\country{China}
}
\email{wutaoliu@nuaa.edu.cn}

\author{YiDan Wang}
\affiliation{%
	\institution{Nanjing University of Aeronautics and Astronautics}
	\city{Nanjing}
	\country{China}
}
\email{wangyidan@nuaa.edu.cn}

\author{Pan Gao}
\affiliation{%
	\institution{Nanjing University of Aeronautics and Astronautics}
	\city{Nanjing}
	\country{China}
}
\email{pan.gao@nuaa.edu.cn}

\renewcommand{\shortauthors}{Trovato et al.}

\begin{abstract}
  Camouflaged object detection (COD) poses a significant challenge in computer vision due to the high similarity between objects and their backgrounds. Existing approaches often rely on heavy training and large computational resources. While foundation models such as the Segment Anything Model (SAM) offer strong generalization, they still struggle to handle COD tasks without fine-tuning and require high-quality prompts to yield good performance. However, generating such prompts manually is costly and inefficient. To address these challenges, we propose \textbf{First RAG, Second SEG (RAG-SEG)}, a training-free paradigm that decouples COD into two stages: Retrieval-Augmented Generation (RAG) for generating coarse masks as prompts, followed by SAM-based segmentation (SEG) for refinement. RAG-SEG constructs a compact retrieval database via unsupervised clustering, enabling fast and effective feature retrieval. During inference, the retrieved features produce pseudo-labels that guide precise mask generation using SAM2. Our method eliminates the need for conventional training while maintaining competitive performance. Extensive experiments on benchmark COD datasets demonstrate that RAG-SEG performs on par with or surpasses state-of-the-art methods. Notably, all experiments are conducted on a \textbf{personal laptop}, highlighting the computational efficiency and practicality of our approach.
	We present further analysis in the Appendix, covering limitations, salient object detection extension, and possible improvements. \textcolor{blue} {Code: \url{https://github.com/Lwt-diamond/RAG-SEG}.}
\end{abstract}





\maketitle

\section{Introduction}
Camouflaged Object Detection (COD) has emerged as a crucial field in both
academic research and industrial applications, with significant implications for
medical imaging analysis, 
wildlife protection, and industrial defect detection. The fundamental
challenge in COD stems from the high visual similarity between target objects
and their environmental context ~\cite{TextureCOD_Galun_Sharon_Basri_Brandt_2003}. 
\begin{figure}
	\centering
	\small
	\includegraphics[width=1\linewidth]{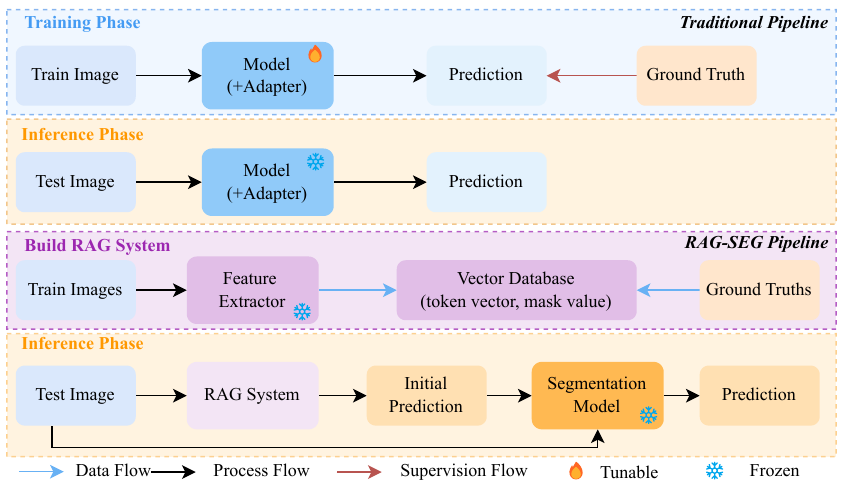}
	\caption{ Comparison between conventional and RAG-SEG approaches for camouflaged object detection. Unlike conventional methods that rely on supervised fine-tuning, RAG-SEG offers a training-free alternative via feature-based retrieval and SAM2-guided refinement. 
		 }
	\label{fig:pipeline_comparison}
\end{figure}
The methodological evolution in COD has progressed from traditional image
processing techniques 
to modern deep
learning approaches. Convolutional Neural Networks (CNNs), particularly SINet~\cite{SINet_fan2020Camouflage},
marked a significant advancement, leading to various CNN-based methods~\cite{SINetv2_fan2021concealed,PFNet_Mei_2021_CVPR,yan2021mirrornet,d2cnet_9430677}.
More recently, attention mechanisms through Vision Transformers (ViT)~\cite{ViT_Dosovitskiy_Beyer_etal._2020}
and transformer-based architectures~\cite{HiNet_aaaihu2022high,HiNet_aaaihu2022high,FSPNet_Huang_Dai_Xiang_Wang_Chen_Qin_Xiong_2023,UGTR_Yang_Zhai_Li_Huang_Luo_Cheng_Fan_2021}
have further enhanced performance. 
\begin{table}[htbp]
	\centering
	\small
	\caption{
		Comparison of computational costs across COD methods.
		The ``Resource'' column lists the GPU model and the corresponding training time (in hours). 
		\textbf{Bold} denotes our method. A hyphen (‘--’) indicates unavailable information.
		(*) For RAG-SEG, the time refers to one-time unsupervised KMeans clustering using FAISS, rather than model training.
	}
	\begin{tabular}{lccc}
		\toprule
		\textbf{Method} & \textbf{Ep.} & \textbf{Resource (Time [h])} & \textbf{Arch.} \\
		\midrule
		SINet$_{20}$~\cite{SINet_fan2020Camouflage}      & 30  & TITAN RTX ×1 (1.17)  & ResNet50~\cite{he2016deep_resnet} \\
		SINetv2$_{22}$~\cite{SINetv2_fan2021concealed}   & 100 & TITAN RTX ×1 (4.00)  & ResNet50~\cite{he2016deep_resnet} \\
		SAM-Adapter$_{23}$~\cite{chen2023samadapter}     & 20  & A100 ×4 (-- )        & SAM-ViT-H~\cite{sam_kirillov2023segany} \\
		DSAM$_{24}$~\cite{dsam_yu2024exploring}          & 100 & 3080Ti ×1 (7.5)      & SAM-ViT-H~\cite{sam_kirillov2023segany} \\
		SAM2-Adapter$_{24}$~\cite{chen2024sam2adapterevaluatingadapting} & 20  & A100 ×3 (-- )        & SAM2~\cite{sam2_ravi2024sam2} \\
		CamoDiff$_{23}$~\cite{CamoDiffchen2023camodiffusion} & 150 & A100 ×1 (-- )        & PVTv2-B4~\cite{wang2021pvtv2_v2} \\
		\textbf{RAG-SEG (Ours)}                         & \textbf{0} & \textbf{Optional (0.13*)} & \textbf{SAM2}~\cite{sam2_ravi2024sam2} \\
		\bottomrule
	\end{tabular}
	\label{tab:method_overview}
\end{table}

 Despite these advances, current COD methods still face critical limitations. Most notably, their reliance on extensive computational resources—often demanding training for over 100 epochs-raises major concerns about \textbf{environmental sustainability and resource efficiency }(Table \ref{tab:method_overview}). In addition, they depend on \textbf{high-end GPUs} to achieve state-of-the-art  performance. 
Recent developments, such as ~\cite{Liu_Zhang_Wu_2022,chen2023samadapter,chen2024sam2adapterevaluatingadapting},
have pushed performance boundaries by implementing a large-scale model with an extremely long
training phase. However, the trend toward increasingly complex models and prolonged training raises concerns about the sustainability and scalability of current approaches.

To address these challenges, we propose \textbf{First RAG, Second SEG (RAG-SEG)}, a novel training-free framework that harnesses off-the-shelf foundation models without requiring extensive training.  
RAG-SEG runs efficiently on standard laptops and eliminates SAM’s dependence on specialized adapters for satisfactory COD performance.
\textbf{Recognizing that SAM's performance hinges critically on prompt quality---and that crafting such prompts manually is both time-consuming and labor-intensive---we integrate Retrieval-Augmented Generation (RAG) to extract key camouflage cues automatically, which then guide SAM's segmentation.}
Figure~\ref{fig:pipeline_comparison} compares the conventional and RAG-SEG methods for COD. We make the following contributions: 
\begin{enumerate}
	\item \textbf{Introduction of RAG-SEG.} We present RAG-SEG, the first of its kind retrieval-augmented generation (RAG) paradigm for object segmentation. By combining RAG’s ability to mine domain-specific cues with the promptable power of SAM, RAG-SEG inaugurates a training-free segmentation pipeline that efficiently harnesses foundation models.
	
	\item \textbf{Training-free COD on a laptop.} We propose a zero-training framework for camouflaged object detection (COD) that runs entirely on an affordable laptop, eliminating the need for GPUs or any traditional training. Extensive experiments on real-world COD benchmarks demonstrate that our method—--despite its zero-training design---matches the performance of existing supervised training approaches.
	
	\item \textbf{Comprehensive empirical validation.} We carry out thorough ablation studies and comparisons against state-of-the-art COD methods, confirming that (i) RAG-SEG’s retrieval components supply SAM with high-quality prompts and (ii) the overall pipeline sustains competitive accuracy with orders-of-magnitude lower resource consumption.	
\end{enumerate}

In addition, we give in-depth analysis of the RAG-SEG pipeline in Appendix, which provides more analysis, extends RAG-SEG to salient object detection, and outlines future research directions.
%
%
%
%

\section{Related Works}
\subsection{Camouflaged Object Detection}

Camouflaged Object Detection (COD) addresses the challenge of identifying
objects that visually assimilate into their surroundings. Traditional approaches
relied on handcrafted features ~\cite{TextureCOD_Galun_Sharon_Basri_Brandt_2003},
which, while pioneering, showed limited effectiveness in complex scenarios
where objects and backgrounds exhibited subtle distinctions. The advent of deep learning, particularly Convolutional Neural Networks (CNNs),
marked a significant advancement in COD. While CNNs enhanced spatial feature extraction
capabilities, they faced limitations in capturing long-range dependencies
crucial for complex camouflage scenarios. This led to the adoption of transformer-based
architectures, which excel at capturing global context but generally demand
substantial computational resources. Recent progress has been driven by comprehensive
datasets like COD10K~\cite{SINet_fan2020Camouflage} and NC4K~\cite{NC4K_Joint_CODSOD_yunqiu_cod21}.
While many methods~\cite{SINet_fan2020Camouflage,SINetv2_fan2021concealed,C2FNet_sun2021c2fnet}
rely on conventional supervised learning approaches, researchers have explored
various complementary features to enhance detection precision. These include
depth information~\cite{Depth-Aided_COD_2023MM,PopNet_wu2023popnet}, frequency
domain features~\cite{FPNet23ACMMM_FrequencuPerception}, 
edge cues~\cite{ji2022fast,FEDER_He2023Camouflaged}, and gradient features~\cite{DGCOD_Ji_Fan_Chou_Dai_Liniger_Gool_2022}.
A comprehensive review can be found in~\cite{CSU_fan2023csu}. While
these improvements have led to better detection accuracy, they also come with notable
trade-offs in terms of computational resources.
\subsection{Retrieval-Augmented Generation}
Retrieval-Augmented Generation (RAG)~\cite{rag_gupta2024comprehensive} has emerged
as an effective approach for mitigating model hallucination in Natural Language
Processing (NLP) and multimodal tasks. \textbf{While RAG has demonstrated
	success in multimodal document retrieval, its potential in segmentation tasks remains
	largely unexplored.} Our work aims to bridge this gap by introducing a novel RAG-based
approach for generating reliable segmentation pseudo-labels.
\subsection{Foundation Models}
Foundation Models (FMs) have transformed the machine learning landscape through
their extensive pre-training on large-scale datasets. While recent COD methods
have explored Parameter-Efficient Fine-Tuning (PEFT~\cite{peft}) approaches
using models like DINOv2~\cite{DINOV2_oquab2023dinov2}, SAM~\cite{sam_kirillov2023segany},
and SAM2~\cite{sam2_ravi2024sam2}, these adaptations often incur significant
computational costs and environmental impact due to gradient storage and training
requirements. In contrast, our proposed RAG-SEG leverages existing FMs
without additional training overhead, enabling efficient deployment even on a laptop. 
\section{Method}
The challenge of COD segmentation stems from the scarcity of labeled datasets and
the high resemblance between foreground and background objects. To achieve
competitive performance in this task, existing approaches rely on large backbones,
extensive parameterization, and long training durations. However, these methods
come at the cost of\textbf{ high computational overheads} and \textbf{ significant environmental
impacts} due to the large amount of energy required for training deep learning models.


Due to the high visual similarity between camouflaged foregrounds and their backgrounds, these off-the-shelf models (SAM~\cite{sam_kirillov2023segany}, SAM2~\cite{sam2_ravi2024sam2}) often fail to produce reliable masks in COD scenarios \cite{chen2023samadapter,chen2024sam2adapterevaluatingadapting}. Rather than designing a separate prompt-tuning network or fine-tuning SAM itself, we propose leveraging Retrieval-Augmented Generation (RAG) \cite{rag_gupta2024comprehensive} to supply SAM with targeted, memory-driven prompts. By maintaining a database of prototype camouflage patterns and using RAG’s retrieve-and-generate to extract salient cue fragments, we can craft context-aware prompts that guide SAM toward accurately delineating camouflaged objects without any additional model training. Our method, RAG-SEG framework, combines the strengths of RAG and SAM, offering a more efficient and carbon-neutral solution for COD segmentation.

\begin{figure*}[htbp]
	\centering
	\includegraphics[width=0.8\textwidth]{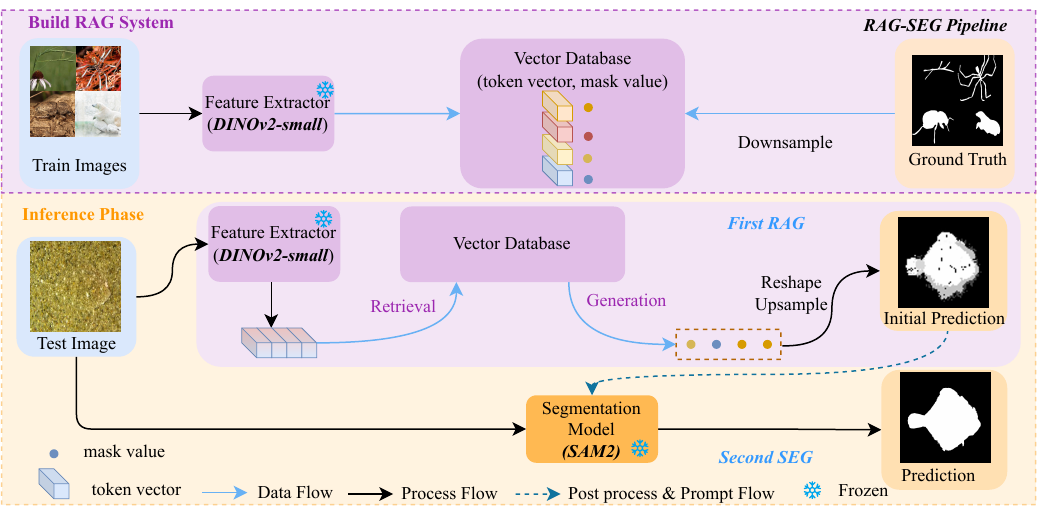} 
	\caption{ Architectural overview of the proposed RAG-SEG, consisting of two stages: (1) RAG system construction via feature extractor and vector database indexing; and (2) inference through retrieval-based mask generation and refinement with SAM2. } 
	\label{fig:ragseg_pipeline}
\end{figure*}


\subsection{Preliminary}
Retrieval-Augmented Generation (RAG)~\cite{rag_gupta2024comprehensive} enhances neural generation by integrating external knowledge retrieval. It comprises a \emph{retriever} and a \emph{generator}: the retriever, often based on dense passage retrieval ~\cite{rag_dpr_karpukhin2020dense}, maps queries into a dense vector space to retrieve relevant context, which the generator then conditions on to produce informed outputs.


In this work, we adapt RAG to improve segmentation performance by
utilizing external knowledge stored in a vector database. This approach
eliminates the need of expensive retraining, providing a more efficient and sustainable
solution.
The overall process follows these key steps:
	\textbf{(1). Vector Database Construction}:
	An embedding model (can be dubbed feature extractor) extracts feature vectors to store in a vector database. In our case, we store the feacture vectors and their corresponding ground truth mask values from training images with DINOv2~\cite{DINOV2_oquab2023dinov2}.
	 \textbf{(2). Query Embedding Generation}: Given a test image, the embedding model (DINOv2) extracts its feature vectors.
	\textbf{(3). Retrieval}:  The system searches the vector database using nearest-neighbor
	algorithms to identify the most relevant entries based on the features.  
	\textbf{(4). Generation}: 
	After obtaining the retrieved vectors and their corresponding mask values, we reshape and upsample them to generate the initial prediction.
	\textbf{(5). Optimization}: 
	Although RAG can yield more accurate results, the initial prediction is still imperfect. It is necessary  to optimize the result using some techniques, like normalization, filtering, and prompt engineering.
\subsection{Overview of Framework}
Directly applying SAM to COD is often suboptimal~\cite{chen2023samadapter,chen2024sam2adapterevaluatingadapting}, since SAM has not been trained on camouflage data and therefore struggles to locate objects that blend into their surroundings. However, SAM possesses a strong ability to interpret well-crafted prompts. We hypothesize that, by supplying SAM with prompts that encode the subtle but essential cues distinguishing camouflaged objects, we can enable it to detect them without any additional training. To realize this training‐free paradigm, we employ a retrieval-augmented generation (RAG) framework to automatically generate such prompts. To the best of our knowledge, this is the first work to design an entirely training‐free method that leverages RAG-generated prompts to adapt SAM for COD.

Our proposed RAG-SEG framework, shown in Figure~\ref{fig:ragseg_pipeline}, introduces a two-stage approach for COD that eliminates the need for traditional training while maintaining competitive segmentation accuracy. After building the vector database,
the final segmentation prediction is obtained through two stages:
\textbf{1. First Stage\textemdash{}RAG}: In this stage, a feature extractor is used to generate
a query from the test image, which is then used to retrieve relevant
information from the vector database. This retrieval process generates initial
predictions with low computational and time complexity. 
\textbf{2. Second Stage\textemdash{}SEG}: The pseudo-label generated in the previous
stage is used as a prompt for the segmentation model, SAM2, after undergoing
post-processing steps such as thresholding. The  segmentation mask is then finally
produced.

\subsection{RAG-based Pseudo-label Generation}
This section presents RAG-based pseudo-label generation in three steps: feature extraction, storage optimization, and retrieval-based generation.

\subsubsection{Feature Extraction}

To construct the vector database, we use the DINOv2~\cite{DINOV2_oquab2023dinov2} Small model as the feature extractor ( \textbf{FE} ). DINOv2 Small is chosen for its efficient performance and relatively small parameter count, offering results comparable to ResNet50~\cite{he2016deep_resnet}, but with superior image representations due to its self-supervised training approach. An alternative, such as using a ResNet-based feature extractor, would generate a \(7 \times 7\)  grid of tokens, resulting in only 49 token-mask pairs per image, which has less image representation. 
Moreover, using ResNet could yield inaccurate downsampled masks, negatively impacting performance. This is why we avoided using a pyramid-style ViT backbone~\cite{liu2021swin, wang2021pyramid_pvt, wang2021pvtv2_v2}.

For each image in the training set, we extract feature vectors from the final
layer of the feature extractor and pair them with corresponding downsampled
mask regions, yielding vector-mask pairs. The input images
are resized to \(224 \times 224\), and with a patch size of 14, the extracted
feature vectors form a \(16 \times 16 \) grid. As a result, each image generates
256 feature vectors, each corresponding to a resized \(16 \times 16 \) mask. For
simplicity and resource efficiency, we omit the use of class tokens. 
The feature vectors and corresponding
mask values are formally represented as follows:
\[
\mathcal{D} =
\left\{
(v_{i}, m_{i}) \ \middle|
 v_{i} = \text{FE}(\mathbf{I}, t_{i}),  m_{i} \in [0, 1],  i \in \{1,2 , \ldots, N\}
\right\}.
\]

Here, \(\mathbf{I}\) denotes the input image, and \(t_{i}\) represents the \(i\)-th
token (patch) of the image. Each \(v_{i}\) is the feature vector corresponding to
the token \(t_{i}\), extracted by the feature extractor. The value \(m_{i}\) is the
corresponding mask value for the token \(t_{i}\), which lies within the range
\([0 , 1]\) due to the downsampling process. Finally, \(N\) is the total number of
tokens (or patches) in the image.
\subsubsection{Optimizing Storage}
Despite the resource efficiency of DINOv2 Small, the size of the resulting vector
database still remains substantial. With 4040 images, and each image generating 256 vector-mask
pairs, the total number of vector-mask pairs can be calculated as: \(4040 \times 256 \approx 10.3424 \times 10^{7} 
\approx  \textbf{103.424 \text{ million}}\). The scale of this dataset presents challenges, particularly in terms of
retrieval time and the computational resources required for subsequent
processing. To mitigate these issues, we apply unsupervised clustering—specifically the KMeans algorithm to compress the database while retaining its representative capacity for segmentation tasks. The clustering process is formalized as follows: 
%
\[
	\mathcal{C} = \text{KMeans}(\mathcal{D}, k=K),
\]
where \(\mathcal{D}\) denotes the original set of vector-mask pairs, \(\mathcal{C}\)
represents the clustered set, and \(K\) is the number of clusters. Experimental results
indicate that setting \(K = 4096 \) achieves a favorable trade-off between storage
reduction and segmentation accuracy.

This clustering approach significantly reduces storage requirements and enhances
retrieval efficiency, enabling faster access to relevant information (vector-mask pairs). The theoretical
basis for this method lies in the observed similarity between adjacent patches
and their corresponding masks within the same image, as well as the similarity
across different images within the same task. 

\subsubsection{Retrieval and Generation}
Given a query image, we extract token-wise features and retrieve the top-$k$ most similar vector-mask pairs from the  database using a similarity function \(f\) (e.g., L2, Inner Product (IP), or Cosine). In practice, IP yields the best results.

Each stored vector \(\mathbf{v}_j\) is assigned a mask value \(m_j \in [0,1]\). For each query token \(\mathbf{q}_i\), its pseudo-label \(\hat{m}_i\) is computed by averaging the mask values of the top-$k$ retrieved vectors from the database:
\[
\hat{m}_i = \tfrac{1}{k} \sum\nolimits_{j=1}^{k} m_j.
\]

Empirically, using $k{=}1$ (i.e., nearest neighbor) achieves the best segmentation performance while simplifying computation.

The final segmentation mask \(\hat{M}_q\) is formed by aggregating pseudo-labels across all token positions:
\[
\hat{M}_q = \{\hat{m}_i \mid i \in [1, N]\}.
\]
This retrieval-based generation produces high-quality pseudo-labels without requiring model fine-tuning, leveraging external knowledge encoded in the feature database.


\subsection{SAM-based Refinement}
\begin{figure}[htpb]
	\centering
	\small
	\includegraphics[width=1\linewidth]{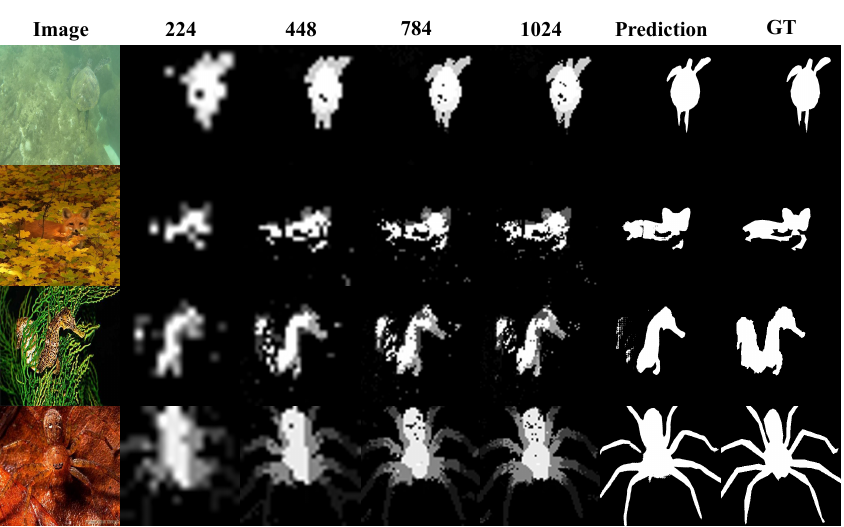}
	\caption{Impact of input resolution on RAG-based detection performance. 
		The two rightmost columns show RAG-SEG results at 784 resolution and ground truth (GT), respectively.
	}
	\label{fig:resolutioncmpv0zip}
\end{figure}

While RAG-based segmentation provides robust coarse localization by exploiting recurring patterns such as occlusion and texture (see Figure~\ref{fig:resolutioncmpv0zip}), its output masks often lack saturation and fine structural detail. Traditional post-processing methods, e.g., Conditional Random Fields (CRF)~\cite{densecrf_krahenbuhl2011efficient}, are ineffective due to the similar appearance statistics of foreground and background in camouflaged scenes. Conversely, off-the-shelf SAM shows limited efficacy in camouflaged object detection without task-specific training. Nonetheless, SAM’s prompt-driven design enables integration of detailed information into segmentation masks. Leveraging these complementary strengths, we employ SAM2 as a refinement module, guided by preliminary RAG-generated masks.

\textbf{Post-Processing Optimization.} To enhance mask prompts for SAM2, we evaluate various thresholding strategies on initial RAG outputs. Experiments indicate a threshold of 0.3 optimally balances segmentation quality and computational efficiency, effectively suppressing noise while preserving structural details to boost SAM2 refinement. Additionally, we explore point prompts to further improve refinement (see Appendix).
\section{Experiments}
\subsection{Datasets and Evaluation }
Following Fan \emph{et al.}~\cite{SINet_fan2020Camouflage}, we employ the training
datasets for camouflaged object detection to validate our method. The training
dataset comprises 4,040 images, with 3,040 images from the COD10K~\cite{SINet_fan2020Camouflage}
dataset and 1,000 images from the CAMO~\cite{AttribbutesCOD_CAMO_Le_Nguyen_Nie_Tran_Sugimoto_2019}
dataset. To evaluate our method, we select the widely-used datasets: COD10K, CAMO,
and CHAMELEON~\cite{CHAMELEON_skurowski2018}. 
To comprehensively evaluate generalization, we adopt multiple metrics capturing complementary aspects. Structure-measure ($S_m$)~\cite{FanStructMeasureICCV17} assesses structural similarity via gradient-based signal comparison. Enhanced alignment with human perception is provided by mean E-measure ($E_{\xi}$)~\cite{Fan2018Enhanced}, which emphasizes fine structures and boundary quality. The weighted F-measure ($F^w_{\beta}$)~\cite{WFMeasureMargolin_Zelnik-Manor_Tal_2014} balances precision and recall through the $\beta$ parameter. Finally, Mean Absolute Error (MAE) measures pixel-wise prediction error, with lower values indicating better accuracy.
%
\subsection{Implementation Details}
Implemented in PyTorch with FAISS~\cite{douze2024faiss}, our framework constructs a feature database from $224 \times 224$ images, clustering features into 4096 centroids via FAISS K-Means (200 iterations; convergence observed at about 150).
    \textbf{Although we do not rely on high-end GPUs, we use an Intel i5-11400H CPU and an NVIDIA GeForce RTX 3050Ti (4GB) on a personal laptop to accelerate feature extraction.}
    For inference,
test images are initially processed at $784 \times 784$ resolution for mask
prediction ($K = 4096$), then downsampled to $256 \times 256$ following the SAM2
pipeline for final segmentation.
\begin{figure*}[htpb]
	\centering
	\small
	\includegraphics[width=0.9\textwidth]{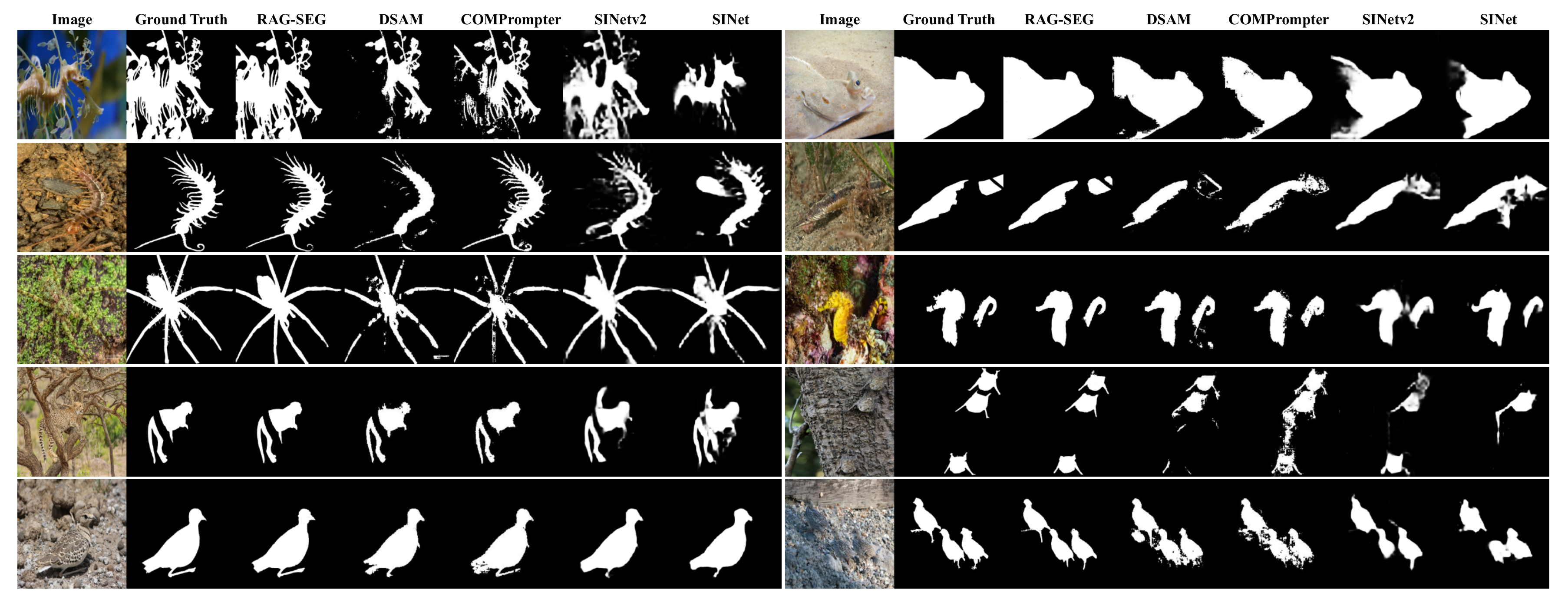}
	
	\caption{Visual comparison of results between our proposed method
		and other SOTA models. 
	}
	\label{fig:ragsegrescmpzip}
\end{figure*}
\subsection{Comparison with State-of-the-Art}
To our knowledge, we are the first to propose a training-free RAG-based method tailored for camouflaged object detection (COD). We compare against several SAM-based COD approaches, including the original SAM~\cite{sam_kirillov2023segany,sam2_ravi2024sam2} and fine-tuned variants~\cite{chen2023samadapter,chen2024sam2adapterevaluatingadapting,comprompter_zhang2024COMPrompter,dsam_yu2024exploring,gensam_hu2024relax,promac_hu2024leveraging}, as well as training-based COD methods~\cite{SINet_fan2020Camouflage,SINetv2_fan2021concealed,PFNet_Mei_2021_CVPR,lin2023frequency}. Table~\ref{tab:quantitative_comparison} summarizes the quantitative comparison across multiple metrics, where arrows indicate whether higher ($\uparrow$) or lower ($\downarrow$) values denote better performance; best results are bolded. As illustrated in Figure~\ref{fig:ragsegrescmpzip}, our method achieves competitive segmentation results, demonstrating robustness in completeness and challenging scenarios. Our qualitative results in Appendix  further highlight its robustness in challenging cases such as \textbf{occlusion}, \textbf{fine-grained structures}, and \textbf{multi-object camouflage}—scenarios that often degrade the performance of even fine-tuned models. This success is attributed to the effectiveness of our initial RAG step, which generates high-quality segmentations capable of capturing occlusions.

\begin{table*}[htpb]
	\centering
	\small
	\caption{Quantitative performance comparison of COD methods. } 
	\begin{tabular}{@{}>{\footnotesize}l@{\hspace{1em}}cccccccccccc@{}}
		\toprule \textbf{Method}                                   & \multicolumn{4}{c}{\textbf{CHAMELEON [10]}} & \multicolumn{4}{c}{\textbf{CAMO [11]}} & \multicolumn{4}{c}{\textbf{COD10K [9]}} \\
		\cmidrule(lr){2-5} \cmidrule(lr){6-9} \cmidrule(lr){10-13} & $S_{\alpha }\uparrow$                       & $E_{\xi}\uparrow$                    & $F_{\beta }^{\omega }\uparrow$         & MAE $\downarrow$ & $S_{\alpha }\uparrow$ &$E_{\xi}\uparrow$ & $F_{\beta }^{\omega }\uparrow$ & MAE $\downarrow$ & $S_{\alpha }\uparrow$ & $E_{\xi}\uparrow$  & $F_{\beta }^{\omega }\uparrow$ & MAE $\downarrow$ \\
		\midrule $\text{SINet}_{2020}$  ~\cite{SINet_fan2020Camouflage}       
		& 0.869 & 0.891 & 0.740 & 0.044 & 0.751 & 0.771 & 0.606 & 0.100 & 0.771 & 0.806 & 0.551 & 0.051 \\
		$\text{SINetv2}_{2021}$ ~\cite{SINetv2_fan2021concealed}     
		& 0.888 & 0.941 & 0.816 & 0.030 & 0.820 & 0.882 & 0.743 & 0.070 & 0.815 & 0.887 & 0.680 & 0.037 \\
		$\text{RankNet}_{2021}$  ~\cite{NC4K_Joint_CODSOD_yunqiu_cod21}     & 0.846 & 0.913 & 0.767 & 0.045 & 0.712 & 0.791 & 0.583 & 0.104 & 0.767 & 0.861 & 0.611 & 0.045 \\
		$\text{SAM-Adapter}_{2023}$ ~\cite{chen2023samadapter} 
		& 0.896 & 0.919 & 0.824 & 0.033 & 0.847 & 0.873 & 0.765 & 0.070 & 0.883 & 0.918 & 0.801 & 0.025 \\
		$\text{SAM2-adapter}_{2024}$ ~\cite{chen2024sam2adapterevaluatingadapting} 
		& \textbf{0.915} & \textbf{0.955} & \textbf{0.889} & \textbf{0.018} 
		& 0.855 & 0.909 & 0.810 & 0.051 
		& \textbf{0.899} & \textbf{0.950} & \textbf{0.850} & \textbf{0.018} \\
		$\text{COMPrompte}_{2024}$ ~\cite{comprompter_zhang2024COMPrompter} 
		& 0.906 & \textbf{0.955} & 0.857 & 0.026 & \textbf{0.882} & \textbf{0.942} & \textbf{0.858} & \textbf{0.044} & 0.889 & 0.949 & 0.821 & 0.023 \\
		$\text{DSAM}_{2024}$  ~\cite{dsam_yu2024exploring}        
		& - & - & - & - & 0.832 & 0.913 & 0.794 & 0.061 & 0.846 & 0.921 & 0.760 & 0.033 \\
        $\text{MDSAM}_{2024}$ ~\cite{mdsam_gao2024multi}
		& - & - & - & - & 0.852 & 0.903 & 0.834 & 0.053 & 0.862 & 0.921 & 0.803 & 0.025 \\
		\midrule
		$\text{SAM}_{2023}$ ~\cite{sam_kirillov2023segany}        
		& 0.727 & 0.734 & 0.639 & 0.081 & 0.684 & 0.687 & 0.606 & 0.132 & 0.783 & 0.798 & 0.701 & 0.050 \\
		$\text{SAM2}_{2024}$   ~\cite{sam2_ravi2024sam2}       
		& 0.359 & 0.375 & 0.115 & 0.357 & 0.350 & 0.411 & 0.079 & 0.311 & 0.429 & 0.505 & 0.115 & 0.218 \\
	   $\text{GenSAM}_{2024}$ ~\cite{gensam_hu2024relax}
		 & 0.774                                       & 0.806                                  & 0.696                                  & 0.073            & 0.729                 & 0.798               & 0.669                          & 0.106            & 0.783                 & 0.843               & 0.695                          & 0.058            \\
	   $\text{ProMaC}_{2024}$~\cite{promac_hu2024leveraging}
		 & 0.833                                       & 0.899                                  & 0.790                                  & 0.044            & 0.767                 & 0.846               & 0.725                          & 0.090            & 0.805                 & 0.876               & 0.716                          & 0.042            \\
		 RAG-SEG     & 0.880 & 0.915 & 0.838 & 0.024 & 0.831 & 0.883 & 0.795 & 0.064 & 0.854 & 0.902 & 0.783 & 0.027 \\
		 
		\bottomrule
	\end{tabular}
	
	\footnotesize
	(Note) The  five rows on the bottom are \textbf{training-free} methods.
	\label{tab:quantitative_comparison}

\end{table*}

\subsection{Ablation Studies}
We conduct comprehensive ablation studies to validate the effectiveness of our
RAG-SEG framework and identify key factors influencing the final segmentation performance.
\textbf{Unless otherwise specified, experiments are performed with $K = 1024$,
	utilizing cosine similarity to retrieve the top-1 most similar token for
	generating the initial prediction on CAMO, with no post-processing
	applied.} 
Additional evaluations—including feature extractor selection, top-\(k\) choice, comparisons with segmentation models such as MobileSAM~\cite{mobile_sam} and SAM~\cite{sam_kirillov2023segany}, the effect of CRF~\cite{densecrf_krahenbuhl2011efficient}, the impact of point prompts and feature extractors, visual comparisons, and extension to salient object detection (SOD)~\cite{duts_wang2017learning,duts_omron_yang2013saliency,ecssd_yan2013hierarchical,pascal_s_li2014secrets,hku_is_li2015visual,mdsam_gao2024multi}—are detailed in the Appendix.

\subsubsection{Impact of Query Image Input Size} 
We employ DINOv2-small for feature extraction, where image resolution significantly
impacts both query generation and processing speed in our RAG-SEG framework.
Experiments on CAMO dataset reveal optimal performance at 784-896 resolution, with
larger sizes offering diminishing returns or even degradation, particularly at
1120-1680 due to feature discrepancies with our 224-resolution training
database (Table~\ref{tab:resolution_comparison}). The qualitative comparison can be seen in Appendix. 
\begin{table}
	\centering
	\small
		\caption{Performance comparison across different query image resolutions on CAMO. } 
	\begin{tabular*}{\linewidth}{@{\extracolsep{\fill}}lcccc@{}}
		\toprule \textbf{Resolution} & $S_{\alpha}\uparrow$ & $F_{\beta }^{\omega }\uparrow$  &  MAE $\downarrow$  & $E_{\xi}\uparrow$     \\
		\midrule 112                 & 0.5775             & 0.3642                       & 0.1375             & 0.5940                 \\
		224                          & 0.6952             & 0.5585                       & 0.1004             & 0.7297                 \\
		448                          & 0.7242             & 0.6163                       & 0.0980             & 0.7694                 \\
		784                          & 0.7369             & \textbf{0.6350}              & \textbf{0.0934}    & 0.7791                 \\
		896                          & \textbf{0.7378}    & 0.6384                       & 0.0942             & \textbf{0.7802}        \\
		1008                         & 0.7286             & 0.6224                       & 0.0969             & 0.7677                 \\
		1024                         & 0.7311             & 0.6272                       & 0.0968             & 0.7699                 \\
		1680                         & 0.6974             & 0.5743                       & 0.1067             & 0.7301                 \\
		\bottomrule
	\end{tabular*}
	\label{tab:resolution_comparison}
\end{table}
\subsubsection{Impact of Thresholding Strategies}
We investigate the effect of different thresholding strategies on the masks generated by RAG. Let the initial probability map be \( P_{\text{initial}} \in [0, 1]^{H \times W} \). We evaluate: (1) no thresholding (\(T_0\), baseline), (2) fixed thresholding at \( n \times 10^{-1} \) (\(T_n\)), and (3) normalized thresholding (\(T_N\)), computed as
\[
P_{\text{norm}} = \frac{P_{\text{initial}} - \min(P_{\text{initial}})}{\max(P_{\text{initial}}) - \min(P_{\text{initial}}) + \epsilon}, \quad \text{where } \epsilon = 10^{-9}.
\]
The experimental results presented in Table~\ref{tab:threshold_impact}
demonstrate several key findings regarding thresholding strategies. First, moderate
thresholding ($T_{3}$) exhibits superior performance across all evaluation metrics,
suggesting an optimal balance in feature selection. Second, aggressive
thresholding ($T_{9}$) leads to a substantial degradation in segmentation
performance, particularly evident in the decreased accuracy metrics. Third, normalized
thresholding ($T_{N}$) demonstrates comparable effectiveness to moderate thresholding
while offering the additional advantage of adaptive scaling capabilities.
These findings indicate that moderate thresholding effectively balances
feature preservation and noise reduction, while excessive thresholding degrades
performance.
\begin{table}
	\centering
	\small
		\caption{Impact of thresholding strategies on segmentation performance on	CAMO.}
	\begin{tabular*}{\linewidth}{@{\extracolsep{\fill}}lcccc@{}}
		\toprule \textbf{Strategy}  & $S_{\alpha}\uparrow$ & $F_{\beta }^{\omega }\uparrow$  &  MAE $\downarrow$  & $E_{\xi}\uparrow$     \\
		\midrule $T_{0}$  & 0.7369             & 0.6350                       & 0.0934             & 0.7791                 \\
		$T_{3}$                     & \textbf{0.7534}    & \textbf{0.6643}              & \textbf{0.0866}    & \textbf{0.7909}        \\
		$T_{5}$                     & 0.7295             & 0.6252                       & 0.0947             & 0.7682                 \\
		$T_{7}$                     & 0.7158             & 0.6000                       & 0.1003             & 0.7538                 \\
		$T_{9}$                     & 0.6883             & 0.5557                       & 0.1077             & 0.7244                 \\
		$T_{N}$        & 0.7371             & 0.6354                       & 0.0933             & 0.7789                 \\
		\bottomrule
	\end{tabular*}
	\label{tab:threshold_impact}
\end{table}
\subsubsection{Impact of Clustering Size} 
\begin{table}
	\centering
	\small
	\caption{
		Comparison of clustering and retrieval times, along with performance metrics, across different cluster sizes.
		“C. Time” is the clustering duration measured on CPU, and “R. Time” is the retrieval latency per \(784\times784\) image.
	}
	\begin{tabular*}{\linewidth}{@{\extracolsep{\fill}}lcccccc@{}}
		\toprule
		\(K\) & C. Time (s) & R. Time (s) & $S_{\alpha}\uparrow$ & $F_{\beta }^{\omega }\uparrow$ & MAE $\downarrow$ & $E_{\xi}\uparrow$ \\
		\midrule
		512   & 4.12   & 0.028731  & 0.7518 & 0.6671 & 0.0893 & 0.7934 \\
		1024  & 14.60  & 0.034090  & 0.7534 & 0.6643 & 0.0866 & 0.7909 \\
		2048  & 57.13  & 0.041342  & \textbf{0.7591} & 0.6755 & 0.0863 & 0.7975 \\
		4096  & 223.00 & 0.055203   & 0.7587 & \textbf{0.6776} & \textbf{0.0858} & \textbf{0.8006} \\
		8192  & 465.42 & 0.088668  & 0.7548 & 0.6709 & 0.0879 & 0.7944 \\
		\bottomrule
	\end{tabular*}
	\label{tab:clustering_comparison} 
\end{table}
Clustering feature vectors by similarity reduces storage demands and accelerates the RAG process. The number of clusters \(K\) in KMeans directly impacts both storage and retrieval efficiency. We evaluate clustering with FAISS under \(K \in [512, 8192]\) using strategy \(T_3\). As shown in Table~\ref{tab:clustering_comparison}, moderate values (2048-4096) yield the best performance: \(K=4096\) achieves optimal $MAE$ (0.0858), $F_{\beta}^{w}$ (0.6776), and $E_{\xi}$ (0.8006), while \(K=2048\) leads in $S_m$ (0.7591).
To assess retrieval efficiency, we measure the time to find top-1 similar vectors under different \(K\). For \(784 \times 784\) images, the number of tokens per query is \(\bigl(\tfrac{784}{14}\bigr)^2\). We adopt the IP metric and include a warm-up phase to avoid cold-start effects. Table~\ref{tab:clustering_comparison} also reports total latency over 1,000 queries and average time per query.

\section{Conclusion}
The RAG-SEG framework demonstrates that effective COD can be achieved without the need of expensive training or huge computational resources. By leveraging RAG to generate initial prompts, our method aids SAM in overcoming the challenges of identifying camouflaged objects. This approach achieves competitive performance while significantly reducing environmental impact and computational requirements. This work lays the foundation for efficient, eco-friendly computer vision in the era of foundation models. Future directions include adapting RAG-SEG to other segmentation tasks (e.g., salient object detection and semantic segmentation), refining the RAG pipeline, exploring adaptive RAG architectures, and developing lightweight SAM alternatives for edge devices.


\bibliographystyle{ACM-Reference-Format}
\bibliography{acmart.bib}


\begin{thebibliography}{52}


\ifx \showCODEN    \undefined \def \showCODEN     #1{\unskip}     \fi
\ifx \showISBNx    \undefined \def \showISBNx     #1{\unskip}     \fi
\ifx \showISBNxiii \undefined \def \showISBNxiii  #1{\unskip}     \fi
\ifx \showISSN     \undefined \def \showISSN      #1{\unskip}     \fi
\ifx \showLCCN     \undefined \def \showLCCN      #1{\unskip}     \fi
\ifx \shownote     \undefined \def \shownote      #1{#1}          \fi
\ifx \showarticletitle \undefined \def \showarticletitle #1{#1}   \fi
\ifx \showURL      \undefined \def \showURL       {\relax}        \fi
\providecommand\bibfield[2]{#2}
\providecommand\bibinfo[2]{#2}
\providecommand\natexlab[1]{#1}
\providecommand\showeprint[2][]{arXiv:#2}

\bibitem[Chen et~al\mbox{.}(2024)]%
        {chen2024sam2adapterevaluatingadapting}
\bibfield{author}{\bibinfo{person}{Tianrun Chen}, \bibinfo{person}{Ankang Lu},
  \bibinfo{person}{Lanyun Zhu}, \bibinfo{person}{Chaotao Ding},
  \bibinfo{person}{Chunan Yu}, \bibinfo{person}{Deyi Ji},
  \bibinfo{person}{Zejian Li}, \bibinfo{person}{Lingyun Sun},
  \bibinfo{person}{Papa Mao}, {and} \bibinfo{person}{Ying Zang}.}
  \bibinfo{year}{2024}\natexlab{}.
\newblock \showarticletitle{Sam2-adapter: Evaluating \& adapting segment
  anything 2 in downstream tasks: Camouflage, shadow, medical image
  segmentation, and more}.
\newblock \bibinfo{journal}{\emph{arXiv preprint arXiv:2408.04579}}
  (\bibinfo{year}{2024}).
\newblock


\bibitem[Chen et~al\mbox{.}(2023b)]%
        {chen2023samadapter}
\bibfield{author}{\bibinfo{person}{Tianrun Chen}, \bibinfo{person}{Lanyun Zhu},
  \bibinfo{person}{Chaotao Deng}, \bibinfo{person}{Runlong Cao},
  \bibinfo{person}{Yan Wang}, \bibinfo{person}{Shangzhan Zhang},
  \bibinfo{person}{Zejian Li}, \bibinfo{person}{Lingyun Sun},
  \bibinfo{person}{Ying Zang}, {and} \bibinfo{person}{Papa Mao}.}
  \bibinfo{year}{2023}\natexlab{b}.
\newblock \showarticletitle{Sam-adapter: Adapting segment anything in
  underperformed scenes}. In \bibinfo{booktitle}{\emph{Proceedings of the
  IEEE/CVF International Conference on Computer Vision}}.
  \bibinfo{pages}{3367--3375}.
\newblock


\bibitem[Chen et~al\mbox{.}(2023a)]%
        {CamoDiffchen2023camodiffusion}
\bibfield{author}{\bibinfo{person}{Zhongxi Chen}, \bibinfo{person}{Ke Sun},
  \bibinfo{person}{Xianming Lin}, {and} \bibinfo{person}{Rongrong Ji}.}
  \bibinfo{year}{2023}\natexlab{a}.
\newblock \showarticletitle{CamoDiffusion: Camouflaged Object Detection via
  Conditional Diffusion Models}.
\newblock \bibinfo{journal}{\emph{arXiv preprint arXiv:2305.17932}}
  (\bibinfo{year}{2023}).
\newblock


\bibitem[Cong et~al\mbox{.}(2023)]%
        {FPNet23ACMMM_FrequencuPerception}
\bibfield{author}{\bibinfo{person}{Runmin Cong}, \bibinfo{person}{Mengyao Sun},
  \bibinfo{person}{Sanyi Zhang}, \bibinfo{person}{Xiaofei Zhou},
  \bibinfo{person}{Wei Zhang}, {and} \bibinfo{person}{Yao Zhao}.}
  \bibinfo{year}{2023}\natexlab{}.
\newblock \showarticletitle{Frequency perception network for camouflaged object
  detection}. In \bibinfo{booktitle}{\emph{Proceedings of the 31st ACM
  International Conference on Multimedia}}. \bibinfo{pages}{1179--1189}.
\newblock


\bibitem[Dosovitskiy et~al\mbox{.}(2021)]%
        {ViT_Dosovitskiy_Beyer_etal._2020}
\bibfield{author}{\bibinfo{person}{Alexey Dosovitskiy}, \bibinfo{person}{Lucas
  Beyer}, \bibinfo{person}{Alexander Kolesnikov}, \bibinfo{person}{Dirk
  Weissenborn}, \bibinfo{person}{Xiaohua Zhai}, \bibinfo{person}{Thomas
  Unterthiner}, \bibinfo{person}{Mostafa Dehghani}, \bibinfo{person}{Matthias
  Minderer}, \bibinfo{person}{Georg Heigold}, \bibinfo{person}{Sylvain Gelly},
  \bibinfo{person}{Jakob Uszkoreit}, {and} \bibinfo{person}{Neil Houlsby}.}
  \bibinfo{year}{2021}\natexlab{}.
\newblock \showarticletitle{An Image is Worth 16x16 Words: Transformers for
  Image Recognition at Scale}.
\newblock  (\bibinfo{year}{2021}).
\newblock
\urldef\tempurl%
\url{https://openreview.net/forum?id=YicbFdNTTy}
\showURL{%
\tempurl}


\bibitem[Douze et~al\mbox{.}(2024)]%
        {douze2024faiss}
\bibfield{author}{\bibinfo{person}{Matthijs Douze}, \bibinfo{person}{Alexandr
  Guzhva}, \bibinfo{person}{Chengqi Deng}, \bibinfo{person}{Jeff Johnson},
  \bibinfo{person}{Gergely Szilvasy}, \bibinfo{person}{Pierre-Emmanuel
  Mazaré}, \bibinfo{person}{Maria Lomeli}, \bibinfo{person}{Lucas Hosseini},
  {and} \bibinfo{person}{Hervé Jégou}.} \bibinfo{year}{2024}\natexlab{}.
\newblock \showarticletitle{The Faiss library}.
\newblock  (\bibinfo{year}{2024}).
\newblock
\showeprint[arxiv]{2401.08281}~[cs.LG]


\bibitem[Fan et~al\mbox{.}(2017)]%
        {FanStructMeasureICCV17}
\bibfield{author}{\bibinfo{person}{Deng-Ping Fan}, \bibinfo{person}{Ming-Ming
  Cheng}, \bibinfo{person}{Yun Liu}, \bibinfo{person}{Tao Li}, {and}
  \bibinfo{person}{Ali Borji}.} \bibinfo{year}{2017}\natexlab{}.
\newblock \showarticletitle{Structure-measure: A New Way to Evaluate Foreground
  Maps}. In \bibinfo{booktitle}{\emph{IEEE International Conference on Computer
  Vision}}.
\newblock


\bibitem[Fan et~al\mbox{.}(2018)]%
        {Fan2018Enhanced}
\bibfield{author}{\bibinfo{person}{Deng-Ping Fan}, \bibinfo{person}{Cheng
  Gong}, \bibinfo{person}{Yang Cao}, \bibinfo{person}{Bo Ren},
  \bibinfo{person}{Ming-Ming Cheng}, {and} \bibinfo{person}{Ali Borji}.}
  \bibinfo{year}{2018}\natexlab{}.
\newblock \showarticletitle{Enhanced-alignment Measure for Binary Foreground
  Map Evaluation}. In \bibinfo{booktitle}{\emph{Proceedings of the
  Twenty-Seventh International Joint Conference on Artificial Intelligence}}.
  AAAI Press.
\newblock


\bibitem[Fan et~al\mbox{.}(2022)]%
        {SINetv2_fan2021concealed}
\bibfield{author}{\bibinfo{person}{Deng-Ping Fan}, \bibinfo{person}{Ge-Peng
  Ji}, \bibinfo{person}{Ming-Ming Cheng}, {and} \bibinfo{person}{Ling Shao}.}
  \bibinfo{year}{2022}\natexlab{}.
\newblock \showarticletitle{Concealed Object Detection}.
\newblock \bibinfo{journal}{\emph{IEEE Transactions on Pattern Analysis and
  Machine Intelligence}} \bibinfo{volume}{44}, \bibinfo{number}{10}
  (\bibinfo{year}{2022}), \bibinfo{pages}{6024--6042}.
\newblock
\href{https://doi.org/10.1109/TPAMI.2021.3085766}{doi:\nolinkurl{10.1109/TPAMI.2021.3085766}}


\bibitem[Fan et~al\mbox{.}(2020)]%
        {SINet_fan2020Camouflage}
\bibfield{author}{\bibinfo{person}{Deng-Ping Fan}, \bibinfo{person}{Ge-Peng
  Ji}, \bibinfo{person}{Guolei Sun}, \bibinfo{person}{Ming-Ming Cheng},
  \bibinfo{person}{Jianbing Shen}, {and} \bibinfo{person}{Ling Shao}.}
  \bibinfo{year}{2020}\natexlab{}.
\newblock \showarticletitle{Camouflaged Object Detection}. In
  \bibinfo{booktitle}{\emph{IEEE Conference on Computer Vision and Pattern
  Recognition (CVPR)}}.
\newblock


\bibitem[Fan et~al\mbox{.}(2023)]%
        {CSU_fan2023csu}
\bibfield{author}{\bibinfo{person}{Deng-Ping Fan}, \bibinfo{person}{Ge-Peng
  Ji}, \bibinfo{person}{Peng Xu}, \bibinfo{person}{Ming-Ming Cheng},
  \bibinfo{person}{Christos Sakaridis}, {and} \bibinfo{person}{Luc Van~Gool}.}
  \bibinfo{year}{2023}\natexlab{}.
\newblock \showarticletitle{Advances in Deep Concealed Scene Understanding}.
\newblock \bibinfo{journal}{\emph{Visual Intelligence (VI)}}
  (\bibinfo{year}{2023}).
\newblock


\bibitem[Galun et~al\mbox{.}(2003)]%
        {TextureCOD_Galun_Sharon_Basri_Brandt_2003}
\bibfield{author}{\bibinfo{person}{Meirav Galun}, \bibinfo{person}{E. Sharon},
  \bibinfo{person}{Ronen Basri}, {and} \bibinfo{person}{Achi Brandt}.}
  \bibinfo{year}{2003}\natexlab{}.
\newblock \showarticletitle{Texture Segmentation by Multiscale Aggregation of
  Filter Responses and Shape Elements}.
\newblock \bibinfo{journal}{\emph{International Conference on Computer Vision}}
  (\bibinfo{date}{Oct} \bibinfo{year}{2003}).
\newblock


\bibitem[Gao et~al\mbox{.}(2024)]%
        {mdsam_gao2024multi}
\bibfield{author}{\bibinfo{person}{Shixuan Gao}, \bibinfo{person}{Pingping
  Zhang}, \bibinfo{person}{Tianyu Yan}, {and} \bibinfo{person}{Huchuan Lu}.}
  \bibinfo{year}{2024}\natexlab{}.
\newblock \showarticletitle{Multi-Scale and Detail-Enhanced Segment Anything
  Model for Salient Object Detection}.
\newblock \bibinfo{journal}{\emph{arXiv preprint arXiv:2408.04326}}
  (\bibinfo{year}{2024}).
\newblock


\bibitem[Gupta et~al\mbox{.}(2024)]%
        {rag_gupta2024comprehensive}
\bibfield{author}{\bibinfo{person}{Shailja Gupta}, \bibinfo{person}{Rajesh
  Ranjan}, {and} \bibinfo{person}{Surya~Narayan Singh}.}
  \bibinfo{year}{2024}\natexlab{}.
\newblock \showarticletitle{A Comprehensive Survey of Retrieval-Augmented
  Generation (RAG): Evolution, Current Landscape and Future Directions}.
\newblock \bibinfo{journal}{\emph{arXiv preprint arXiv:2410.12837}}
  (\bibinfo{year}{2024}).
\newblock


\bibitem[He et~al\mbox{.}(2023)]%
        {FEDER_He2023Camouflaged}
\bibfield{author}{\bibinfo{person}{Chunming He}, \bibinfo{person}{Kai Li},
  \bibinfo{person}{Yachao Zhang}, \bibinfo{person}{Longxiang Tang},
  \bibinfo{person}{Yulun Zhang}, \bibinfo{person}{Zhenhua Guo}, {and}
  \bibinfo{person}{Xiu Li}.} \bibinfo{year}{2023}\natexlab{}.
\newblock \showarticletitle{Camouflaged Object Detection with Feature
  Decomposition and Edge Reconstruction}. In \bibinfo{booktitle}{\emph{IEEE
  Conference on Computer Vision and Pattern Recognition (CVPR)}}.
\newblock


\bibitem[He et~al\mbox{.}(2016)]%
        {he2016deep_resnet}
\bibfield{author}{\bibinfo{person}{Kaiming He}, \bibinfo{person}{Xiangyu
  Zhang}, \bibinfo{person}{Shaoqing Ren}, {and} \bibinfo{person}{Jian Sun}.}
  \bibinfo{year}{2016}\natexlab{}.
\newblock \showarticletitle{Deep residual learning for image recognition}. In
  \bibinfo{booktitle}{\emph{Proceedings of the IEEE conference on computer
  vision and pattern recognition}}. \bibinfo{pages}{770--778}.
\newblock


\bibitem[Hu et~al\mbox{.}(2024a)]%
        {gensam_hu2024relax}
\bibfield{author}{\bibinfo{person}{Jian Hu}, \bibinfo{person}{Jiayi Lin},
  \bibinfo{person}{Shaogang Gong}, {and} \bibinfo{person}{Weitong Cai}.}
  \bibinfo{year}{2024}\natexlab{a}.
\newblock \showarticletitle{Relax Image-Specific Prompt Requirement in SAM: A
  Single Generic Prompt for Segmenting Camouflaged Objects}. In
  \bibinfo{booktitle}{\emph{Proceedings of the AAAI Conference on Artificial
  Intelligence}}, Vol.~\bibinfo{volume}{38}. \bibinfo{pages}{12511--12518}.
\newblock


\bibitem[Hu et~al\mbox{.}(2024b)]%
        {promac_hu2024leveraging}
\bibfield{author}{\bibinfo{person}{Jian Hu}, \bibinfo{person}{Jiayi Lin},
  \bibinfo{person}{Junchi Yan}, {and} \bibinfo{person}{Shaogang Gong}.}
  \bibinfo{year}{2024}\natexlab{b}.
\newblock \showarticletitle{Leveraging Hallucinations to Reduce Manual Prompt
  Dependency in Promptable Segmentation}.
\newblock \bibinfo{journal}{\emph{arXiv preprint arXiv:2408.15205}}
  (\bibinfo{year}{2024}).
\newblock


\bibitem[Hu et~al\mbox{.}(2022)]%
        {HiNet_aaaihu2022high}
\bibfield{author}{\bibinfo{person}{Xiaobin Hu}, \bibinfo{person}{Shuo Wang},
  \bibinfo{person}{Xuebin Qin}, \bibinfo{person}{Hang Dai},
  \bibinfo{person}{Wenqi Ren}, \bibinfo{person}{Ying Tai},
  \bibinfo{person}{Chengjie Wang}, {and} \bibinfo{person}{Ling Shao}.}
  \bibinfo{year}{2022}\natexlab{}.
\newblock \showarticletitle{High-resolution Iterative Feedback Network for
  Camouflaged Object Detection}.
\newblock \bibinfo{journal}{\emph{arXiv preprint arXiv:2203.11624}}
  (\bibinfo{year}{2022}).
\newblock


\bibitem[Huang et~al\mbox{.}(2023)]%
        {FSPNet_Huang_Dai_Xiang_Wang_Chen_Qin_Xiong_2023}
\bibfield{author}{\bibinfo{person}{Zhou Huang}, \bibinfo{person}{Hang Dai},
  \bibinfo{person}{Tian-Zhu Xiang}, \bibinfo{person}{Shuo Wang},
  \bibinfo{person}{Huai-Xin Chen}, \bibinfo{person}{Jie Qin}, {and}
  \bibinfo{person}{Huan Xiong}.} \bibinfo{year}{2023}\natexlab{}.
\newblock \showarticletitle{Feature Shrinkage Pyramid for Camouflaged Object
  Detection with Transformers}.
\newblock  (\bibinfo{year}{2023}).
\newblock


\bibitem[Ji et~al\mbox{.}(2023)]%
        {DGCOD_Ji_Fan_Chou_Dai_Liniger_Gool_2022}
\bibfield{author}{\bibinfo{person}{Ge-Peng Ji}, \bibinfo{person}{Deng-Ping
  Fan}, \bibinfo{person}{Yu-Cheng Chou}, \bibinfo{person}{Dengxin Dai},
  \bibinfo{person}{Alexander Liniger}, {and} \bibinfo{person}{Luc Van~Gool}.}
  \bibinfo{year}{2023}\natexlab{}.
\newblock \showarticletitle{Deep Gradient Learning for Efficient Camouflaged
  Object Detection}.
\newblock \bibinfo{journal}{\emph{Machine Intelligence Research}}
  \bibinfo{volume}{20} (\bibinfo{year}{2023}), \bibinfo{pages}{92--108}.
\newblock
Issue 1.


\bibitem[Ji et~al\mbox{.}(2022)]%
        {ji2022fast}
\bibfield{author}{\bibinfo{person}{Ge-Peng Ji}, \bibinfo{person}{Lei Zhu},
  \bibinfo{person}{Mingchen Zhuge}, {and} \bibinfo{person}{Keren Fu}.}
  \bibinfo{year}{2022}\natexlab{}.
\newblock \showarticletitle{Fast camouflaged object detection via edge-based
  reversible re-calibration network}.
\newblock \bibinfo{journal}{\emph{Pattern Recognition}}  \bibinfo{volume}{123}
  (\bibinfo{year}{2022}), \bibinfo{pages}{108414}.
\newblock


\bibitem[Karpukhin et~al\mbox{.}(2020)]%
        {rag_dpr_karpukhin2020dense}
\bibfield{author}{\bibinfo{person}{Vladimir Karpukhin}, \bibinfo{person}{Barlas
  O{\u{g}}uz}, \bibinfo{person}{Sewon Min}, \bibinfo{person}{Patrick Lewis},
  \bibinfo{person}{Ledell Wu}, \bibinfo{person}{Sergey Edunov},
  \bibinfo{person}{Danqi Chen}, {and} \bibinfo{person}{Wen-tau Yih}.}
  \bibinfo{year}{2020}\natexlab{}.
\newblock \showarticletitle{Dense passage retrieval for open-domain question
  answering}.
\newblock \bibinfo{journal}{\emph{arXiv preprint arXiv:2004.04906}}
  (\bibinfo{year}{2020}).
\newblock


\bibitem[Kirillov et~al\mbox{.}(2023)]%
        {sam_kirillov2023segany}
\bibfield{author}{\bibinfo{person}{Alexander Kirillov}, \bibinfo{person}{Eric
  Mintun}, \bibinfo{person}{Nikhila Ravi}, \bibinfo{person}{Hanzi Mao},
  \bibinfo{person}{Chloe Rolland}, \bibinfo{person}{Laura Gustafson},
  \bibinfo{person}{Tete Xiao}, \bibinfo{person}{Spencer Whitehead},
  \bibinfo{person}{Alexander~C. Berg}, \bibinfo{person}{Wan-Yen Lo},
  \bibinfo{person}{Piotr Doll{\'a}r}, {and} \bibinfo{person}{Ross Girshick}.}
  \bibinfo{year}{2023}\natexlab{}.
\newblock \showarticletitle{Segment Anything}.
\newblock \bibinfo{journal}{\emph{arXiv:2304.02643}} (\bibinfo{year}{2023}).
\newblock


\bibitem[Kr{\"a}henb{\"u}hl and Koltun(2011)]%
        {densecrf_krahenbuhl2011efficient}
\bibfield{author}{\bibinfo{person}{Philipp Kr{\"a}henb{\"u}hl} {and}
  \bibinfo{person}{Vladlen Koltun}.} \bibinfo{year}{2011}\natexlab{}.
\newblock \showarticletitle{Efficient inference in fully connected crfs with
  gaussian edge potentials}.
\newblock \bibinfo{journal}{\emph{Advances in neural information processing
  systems}}  \bibinfo{volume}{24} (\bibinfo{year}{2011}).
\newblock


\bibitem[Le et~al\mbox{.}(2019)]%
        {AttribbutesCOD_CAMO_Le_Nguyen_Nie_Tran_Sugimoto_2019}
\bibfield{author}{\bibinfo{person}{Trung-Nghia Le}, \bibinfo{person}{Tam~V.
  Nguyen}, \bibinfo{person}{Zhongliang Nie}, \bibinfo{person}{Minh-Triet Tran},
  {and} \bibinfo{person}{Akihiro Sugimoto}.} \bibinfo{year}{2019}\natexlab{}.
\newblock \showarticletitle{Anabranch network for camouflaged object
  segmentation}.
\newblock \bibinfo{journal}{\emph{Computer Vision and Image Understanding}}
  (\bibinfo{date}{Jul} \bibinfo{year}{2019}), \bibinfo{pages}{45–56}.
\newblock
\href{https://doi.org/10.1016/j.cviu.2019.04.006}{doi:\nolinkurl{10.1016/j.cviu.2019.04.006}}


\bibitem[Li and Yu(2015)]%
        {hku_is_li2015visual}
\bibfield{author}{\bibinfo{person}{Guanbin Li} {and} \bibinfo{person}{Yizhou
  Yu}.} \bibinfo{year}{2015}\natexlab{}.
\newblock \showarticletitle{Visual saliency based on multiscale deep features}.
  In \bibinfo{booktitle}{\emph{Proceedings of the IEEE conference on computer
  vision and pattern recognition}}. \bibinfo{pages}{5455--5463}.
\newblock


\bibitem[Li et~al\mbox{.}(2014)]%
        {pascal_s_li2014secrets}
\bibfield{author}{\bibinfo{person}{Yin Li}, \bibinfo{person}{Xiaodi Hou},
  \bibinfo{person}{Christof Koch}, \bibinfo{person}{James~M Rehg}, {and}
  \bibinfo{person}{Alan~L Yuille}.} \bibinfo{year}{2014}\natexlab{}.
\newblock \showarticletitle{The secrets of salient object segmentation}. In
  \bibinfo{booktitle}{\emph{Proceedings of the IEEE conference on computer
  vision and pattern recognition}}. \bibinfo{pages}{280--287}.
\newblock


\bibitem[Lin et~al\mbox{.}(2023)]%
        {lin2023frequency}
\bibfield{author}{\bibinfo{person}{Jiaying Lin}, \bibinfo{person}{Xin Tan},
  \bibinfo{person}{Ke Xu}, \bibinfo{person}{Lizhuang Ma}, {and}
  \bibinfo{person}{Rynson~WH Lau}.} \bibinfo{year}{2023}\natexlab{}.
\newblock \showarticletitle{Frequency-aware camouflaged object detection}.
\newblock \bibinfo{journal}{\emph{ACM Transactions on Multimedia Computing,
  Communications and Applications}} \bibinfo{volume}{19}, \bibinfo{number}{2}
  (\bibinfo{year}{2023}), \bibinfo{pages}{1--16}.
\newblock


\bibitem[Liu et~al\mbox{.}(2021)]%
        {liu2021swin}
\bibfield{author}{\bibinfo{person}{Ze Liu}, \bibinfo{person}{Yutong Lin},
  \bibinfo{person}{Yue Cao}, \bibinfo{person}{Han Hu}, \bibinfo{person}{Yixuan
  Wei}, \bibinfo{person}{Zheng Zhang}, \bibinfo{person}{Stephen Lin}, {and}
  \bibinfo{person}{Baining Guo}.} \bibinfo{year}{2021}\natexlab{}.
\newblock \showarticletitle{Swin transformer: Hierarchical vision transformer
  using shifted windows}. In \bibinfo{booktitle}{\emph{Proceedings of the
  IEEE/CVF international conference on computer vision}}.
  \bibinfo{pages}{10012--10022}.
\newblock


\bibitem[Liu et~al\mbox{.}(2022)]%
        {Liu_Zhang_Wu_2022}
\bibfield{author}{\bibinfo{person}{Zhengyi Liu}, \bibinfo{person}{Zhili Zhang},
  \bibinfo{person}{Yacheng Tan}, {and} \bibinfo{person}{Wei Wu}.}
  \bibinfo{year}{2022}\natexlab{}.
\newblock \showarticletitle{Boosting Camouflaged Object Detection with
  Dual-Task Interactive Transformer}.
\newblock  (\bibinfo{year}{2022}), \bibinfo{pages}{140--146}.
\newblock


\bibitem[Lyu et~al\mbox{.}(2021)]%
        {NC4K_Joint_CODSOD_yunqiu_cod21}
\bibfield{author}{\bibinfo{person}{Yunqiu Lyu}, \bibinfo{person}{Jing Zhang},
  \bibinfo{person}{Yuchao Dai}, \bibinfo{person}{Aixuan Li},
  \bibinfo{person}{Bowen Liu}, \bibinfo{person}{Nick Barnes}, {and}
  \bibinfo{person}{Deng-Ping Fan}.} \bibinfo{year}{2021}\natexlab{}.
\newblock \showarticletitle{Simultaneously Localize, Segment and Rank the
  Camouflaged Objects}. In \bibinfo{booktitle}{\emph{Proceedings of the
  IEEE/CVF Conference on Computer Vision and Pattern Recognition (CVPR)}}.
\newblock


\bibitem[Mangrulkar et~al\mbox{.}(2022)]%
        {peft}
\bibfield{author}{\bibinfo{person}{Sourab Mangrulkar}, \bibinfo{person}{Sylvain
  Gugger}, \bibinfo{person}{Lysandre Debut}, \bibinfo{person}{Younes Belkada},
  \bibinfo{person}{Sayak Paul}, {and} \bibinfo{person}{Benjamin Bossan}.}
  \bibinfo{year}{2022}\natexlab{}.
\newblock \bibinfo{title}{PEFT: State-of-the-art Parameter-Efficient
  Fine-Tuning methods}.
\newblock \bibinfo{howpublished}{\url{https://github.com/huggingface/peft}}.
\newblock


\bibitem[Margolin et~al\mbox{.}(2014)]%
        {WFMeasureMargolin_Zelnik-Manor_Tal_2014}
\bibfield{author}{\bibinfo{person}{Ran Margolin}, \bibinfo{person}{Lihi
  Zelnik-Manor}, {and} \bibinfo{person}{Ayellet Tal}.}
  \bibinfo{year}{2014}\natexlab{}.
\newblock \showarticletitle{How to Evaluate Foreground Maps}. In
  \bibinfo{booktitle}{\emph{2014 IEEE Conference on Computer Vision and Pattern
  Recognition}}.
\newblock
\href{https://doi.org/10.1109/cvpr.2014.39}{doi:\nolinkurl{10.1109/cvpr.2014.39}}


\bibitem[Mei et~al\mbox{.}(2021)]%
        {PFNet_Mei_2021_CVPR}
\bibfield{author}{\bibinfo{person}{Haiyang Mei}, \bibinfo{person}{Ge-Peng Ji},
  \bibinfo{person}{Ziqi Wei}, \bibinfo{person}{Xin Yang},
  \bibinfo{person}{Xiaopeng Wei}, {and} \bibinfo{person}{Deng-Ping Fan}.}
  \bibinfo{year}{2021}\natexlab{}.
\newblock \showarticletitle{Camouflaged Object Segmentation with Distraction
  Mining}. In \bibinfo{booktitle}{\emph{IEEE/CVF Conference on Computer Vision
  and Pattern Recognition (CVPR)}}.
\newblock


\bibitem[Oquab et~al\mbox{.}(2023)]%
        {DINOV2_oquab2023dinov2}
\bibfield{author}{\bibinfo{person}{Maxime Oquab}, \bibinfo{person}{Timoth{\'e}e
  Darcet}, \bibinfo{person}{Th{\'e}o Moutakanni}, \bibinfo{person}{Huy Vo},
  \bibinfo{person}{Marc Szafraniec}, \bibinfo{person}{Vasil Khalidov},
  \bibinfo{person}{Pierre Fernandez}, \bibinfo{person}{Daniel Haziza},
  \bibinfo{person}{Francisco Massa}, \bibinfo{person}{Alaaeldin El-Nouby},
  {et~al\mbox{.}}} \bibinfo{year}{2023}\natexlab{}.
\newblock \bibinfo{title}{Dinov2: Learning robust visual features without
  supervision}.
\newblock


\bibitem[Przemysław~Skurowski and Kornacki(2018)]%
        {CHAMELEON_skurowski2018}
\bibfield{author}{\bibinfo{person}{Jakub Błaszczyk Tomasz~Depta
  Przemysław~Skurowski, Hassan~Abdulameer} {and} \bibinfo{person}{Adam
  Kornacki}.} \bibinfo{year}{2018}\natexlab{}.
\newblock \showarticletitle{Animal camouflage analysis: Chameleon database}.
\newblock \bibinfo{journal}{\emph{Unpublished manuscript}} \bibinfo{volume}{2},
  \bibinfo{number}{6} (\bibinfo{year}{2018}), \bibinfo{pages}{7}.
\newblock


\bibitem[Ravi et~al\mbox{.}(2024)]%
        {sam2_ravi2024sam2}
\bibfield{author}{\bibinfo{person}{Nikhila Ravi}, \bibinfo{person}{Valentin
  Gabeur}, \bibinfo{person}{Yuan-Ting Hu}, \bibinfo{person}{Ronghang Hu},
  \bibinfo{person}{Chaitanya Ryali}, \bibinfo{person}{Tengyu Ma},
  \bibinfo{person}{Haitham Khedr}, \bibinfo{person}{Roman R{\"a}dle},
  \bibinfo{person}{Chloe Rolland}, \bibinfo{person}{Laura Gustafson},
  \bibinfo{person}{Eric Mintun}, \bibinfo{person}{Junting Pan},
  \bibinfo{person}{Kalyan~Vasudev Alwala}, \bibinfo{person}{Nicolas Carion},
  \bibinfo{person}{Chao-Yuan Wu}, \bibinfo{person}{Ross Girshick},
  \bibinfo{person}{Piotr Doll{\'a}r}, {and} \bibinfo{person}{Christoph
  Feichtenhofer}.} \bibinfo{year}{2024}\natexlab{}.
\newblock \showarticletitle{SAM 2: Segment Anything in Images and Videos}.
\newblock \bibinfo{journal}{\emph{arXiv preprint arXiv:2408.00714}}
  (\bibinfo{year}{2024}).
\newblock
\urldef\tempurl%
\url{https://arxiv.org/abs/2408.00714}
\showURL{%
\tempurl}


\bibitem[Sun et~al\mbox{.}(2021)]%
        {C2FNet_sun2021c2fnet}
\bibfield{author}{\bibinfo{person}{Yujia Sun}, \bibinfo{person}{Geng Chen},
  \bibinfo{person}{Tao Zhou}, \bibinfo{person}{Yi Zhang}, {and}
  \bibinfo{person}{Nian Liu}.} \bibinfo{year}{2021}\natexlab{}.
\newblock \showarticletitle{Context-aware Cross-level Fusion Network for
  Camouflaged Object Detection}. In \bibinfo{booktitle}{\emph{Proceedings of
  the 30th International Joint Conference on Artificial Intelligence}}.
  \bibinfo{pages}{1025--1031}.
\newblock


\bibitem[Wang et~al\mbox{.}(2022a)]%
        {d2cnet_9430677}
\bibfield{author}{\bibinfo{person}{Kang Wang}, \bibinfo{person}{Hongbo Bi},
  \bibinfo{person}{Yi Zhang}, \bibinfo{person}{Cong Zhang},
  \bibinfo{person}{Ziqi Liu}, {and} \bibinfo{person}{Shuang Zheng}.}
  \bibinfo{year}{2022}\natexlab{a}.
\newblock \showarticletitle{D$^{2}$C-Net: A Dual-Branch, Dual-Guidance and
  Cross-Refine Network for Camouflaged Object Detection}.
\newblock \bibinfo{journal}{\emph{IEEE Transactions on Industrial Electronics}}
  \bibinfo{volume}{69}, \bibinfo{number}{5} (\bibinfo{year}{2022}),
  \bibinfo{pages}{5364--5374}.
\newblock
\href{https://doi.org/10.1109/TIE.2021.3078379}{doi:\nolinkurl{10.1109/TIE.2021.3078379}}


\bibitem[Wang et~al\mbox{.}(2017)]%
        {duts_wang2017learning}
\bibfield{author}{\bibinfo{person}{Lijun Wang}, \bibinfo{person}{Huchuan Lu},
  \bibinfo{person}{Yifan Wang}, \bibinfo{person}{Mengyang Feng},
  \bibinfo{person}{Dong Wang}, \bibinfo{person}{Baocai Yin}, {and}
  \bibinfo{person}{Xiang Ruan}.} \bibinfo{year}{2017}\natexlab{}.
\newblock \showarticletitle{Learning to detect salient objects with image-level
  supervision}. In \bibinfo{booktitle}{\emph{Proceedings of the IEEE conference
  on computer vision and pattern recognition}}. \bibinfo{pages}{136--145}.
\newblock


\bibitem[Wang et~al\mbox{.}(2023)]%
        {Depth-Aided_COD_2023MM}
\bibfield{author}{\bibinfo{person}{Qingwei Wang}, \bibinfo{person}{Jinyu Yang},
  \bibinfo{person}{Xiaosheng Yu}, \bibinfo{person}{Fangyi Wang},
  \bibinfo{person}{Peng Chen}, {and} \bibinfo{person}{Feng Zheng}.}
  \bibinfo{year}{2023}\natexlab{}.
\newblock \showarticletitle{Depth-Aided Camouflaged Object Detection}. In
  \bibinfo{booktitle}{\emph{Proceedings of the 31st ACM International
  Conference on Multimedia}} \emph{(\bibinfo{series}{MM '23})}.
  \bibinfo{publisher}{Association for Computing Machinery}.
\newblock


\bibitem[Wang et~al\mbox{.}(2021)]%
        {wang2021pyramid_pvt}
\bibfield{author}{\bibinfo{person}{Wenhai Wang}, \bibinfo{person}{Enze Xie},
  \bibinfo{person}{Xiang Li}, \bibinfo{person}{Deng-Ping Fan},
  \bibinfo{person}{Kaitao Song}, \bibinfo{person}{Ding Liang},
  \bibinfo{person}{Tong Lu}, \bibinfo{person}{Ping Luo}, {and}
  \bibinfo{person}{Ling Shao}.} \bibinfo{year}{2021}\natexlab{}.
\newblock \showarticletitle{Pyramid vision transformer: A versatile backbone
  for dense prediction without convolutions}. In
  \bibinfo{booktitle}{\emph{Proceedings of the IEEE/CVF International
  Conference on Computer Vision}}. \bibinfo{pages}{568--578}.
\newblock


\bibitem[Wang et~al\mbox{.}(2022b)]%
        {wang2021pvtv2_v2}
\bibfield{author}{\bibinfo{person}{Wenhai Wang}, \bibinfo{person}{Enze Xie},
  \bibinfo{person}{Xiang Li}, \bibinfo{person}{Deng-Ping Fan},
  \bibinfo{person}{Kaitao Song}, \bibinfo{person}{Ding Liang},
  \bibinfo{person}{Tong Lu}, \bibinfo{person}{Ping Luo}, {and}
  \bibinfo{person}{Ling Shao}.} \bibinfo{year}{2022}\natexlab{b}.
\newblock \showarticletitle{Pvtv2: Improved baselines with pyramid vision
  transformer}.
\newblock \bibinfo{journal}{\emph{Computational Visual Media}}
  \bibinfo{volume}{8}, \bibinfo{number}{3} (\bibinfo{year}{2022}),
  \bibinfo{pages}{1--10}.
\newblock


\bibitem[Wu et~al\mbox{.}(2023)]%
        {PopNet_wu2023popnet}
\bibfield{author}{\bibinfo{person}{Zongwei Wu}, \bibinfo{person}{Danda~Pani
  Paudel}, \bibinfo{person}{Deng-Ping Fan}, \bibinfo{person}{Jingjing Wang},
  \bibinfo{person}{Shuo Wang}, \bibinfo{person}{Cédric Demonceaux},
  \bibinfo{person}{Radu Timofte}, {and} \bibinfo{person}{Luc Van~Gool}.}
  \bibinfo{year}{2023}\natexlab{}.
\newblock \showarticletitle{Source-free depth for object pop-out}. In
  \bibinfo{booktitle}{\emph{Proceedings of the IEEE/CVF International
  Conference on Computer Vision (ICCV)}}.
\newblock


\bibitem[Yan et~al\mbox{.}(2021)]%
        {yan2021mirrornet}
\bibfield{author}{\bibinfo{person}{Jinnan Yan}, \bibinfo{person}{Trung-Nghia
  Le}, \bibinfo{person}{Khanh-Duy Nguyen}, \bibinfo{person}{Minh-Triet Tran},
  \bibinfo{person}{Thanh-Toan Do}, {and} \bibinfo{person}{Tam~V Nguyen}.}
  \bibinfo{year}{2021}\natexlab{}.
\newblock \showarticletitle{Mirrornet: Bio-inspired camouflaged object
  segmentation}.
\newblock \bibinfo{journal}{\emph{IEEE Access}}  \bibinfo{volume}{9}
  (\bibinfo{year}{2021}), \bibinfo{pages}{43290--43300}.
\newblock


\bibitem[Yan et~al\mbox{.}(2013)]%
        {ecssd_yan2013hierarchical}
\bibfield{author}{\bibinfo{person}{Qiong Yan}, \bibinfo{person}{Li Xu},
  \bibinfo{person}{Jianping Shi}, {and} \bibinfo{person}{Jiaya Jia}.}
  \bibinfo{year}{2013}\natexlab{}.
\newblock \showarticletitle{Hierarchical saliency detection}. In
  \bibinfo{booktitle}{\emph{Proceedings of the IEEE conference on computer
  vision and pattern recognition}}. \bibinfo{pages}{1155--1162}.
\newblock


\bibitem[Yang et~al\mbox{.}(2013)]%
        {duts_omron_yang2013saliency}
\bibfield{author}{\bibinfo{person}{Chuan Yang}, \bibinfo{person}{Lihe Zhang},
  \bibinfo{person}{Huchuan Lu}, \bibinfo{person}{Xiang Ruan}, {and}
  \bibinfo{person}{Ming-Hsuan Yang}.} \bibinfo{year}{2013}\natexlab{}.
\newblock \showarticletitle{Saliency detection via graph-based manifold
  ranking}. In \bibinfo{booktitle}{\emph{Proceedings of the IEEE conference on
  computer vision and pattern recognition}}. \bibinfo{pages}{3166--3173}.
\newblock


\bibitem[Yang et~al\mbox{.}(2021)]%
        {UGTR_Yang_Zhai_Li_Huang_Luo_Cheng_Fan_2021}
\bibfield{author}{\bibinfo{person}{Fan Yang}, \bibinfo{person}{Qiang Zhai},
  \bibinfo{person}{Xin Li}, \bibinfo{person}{Rui Huang}, \bibinfo{person}{Ao
  Luo}, \bibinfo{person}{Hong Cheng}, {and} \bibinfo{person}{Deng-Ping Fan}.}
  \bibinfo{year}{2021}\natexlab{}.
\newblock \showarticletitle{Uncertainty-Guided Transformer Reasoning for
  Camouflaged Object Detection}. In \bibinfo{booktitle}{\emph{2021 IEEE/CVF
  International Conference on Computer Vision (ICCV)}}.
\newblock
\href{https://doi.org/10.1109/iccv48922.2021.00411}{doi:\nolinkurl{10.1109/iccv48922.2021.00411}}


\bibitem[Yu et~al\mbox{.}(2024)]%
        {dsam_yu2024exploring}
\bibfield{author}{\bibinfo{person}{Zhenni Yu}, \bibinfo{person}{Xiaoqin Zhang},
  \bibinfo{person}{Li Zhao}, \bibinfo{person}{Yi Bin}, {and}
  \bibinfo{person}{Guobao Xiao}.} \bibinfo{year}{2024}\natexlab{}.
\newblock \showarticletitle{Exploring Deeper! Segment Anything Model with Depth
  Perception for Camouflaged Object Detection}. In
  \bibinfo{booktitle}{\emph{Proceedings of the 32nd ACM International
  Conference on Multimedia (ACM MM 2024)}}. \bibinfo{publisher}{Association for
  Computing Machinery}, \bibinfo{pages}{123--132}.
\newblock


\bibitem[Zhang et~al\mbox{.}(2023)]%
        {mobile_sam}
\bibfield{author}{\bibinfo{person}{Chaoning Zhang}, \bibinfo{person}{Dongshen
  Han}, \bibinfo{person}{Yu Qiao}, \bibinfo{person}{Jung~Uk Kim},
  \bibinfo{person}{Sung-Ho Bae}, \bibinfo{person}{Seungkyu Lee}, {and}
  \bibinfo{person}{Choong~Seon Hong}.} \bibinfo{year}{2023}\natexlab{}.
\newblock \showarticletitle{Faster Segment Anything: Towards Lightweight SAM
  for Mobile Applications}.
\newblock \bibinfo{journal}{\emph{arXiv preprint arXiv:2306.14289}}
  (\bibinfo{year}{2023}).
\newblock


\bibitem[Zhang et~al\mbox{.}(2024)]%
        {comprompter_zhang2024COMPrompter}
\bibfield{author}{\bibinfo{person}{Xiaoqin Zhang}, \bibinfo{person}{Zhenni Yu},
  \bibinfo{person}{Li Zhao}, \bibinfo{person}{Deng-Ping Fan}, {and}
  \bibinfo{person}{Guobao Xiao}.} \bibinfo{year}{2024}\natexlab{}.
\newblock \showarticletitle{COMPrompter: Rethink SAM in Camouflaged Object
  Detection with Multi-Prompt Network}.
\newblock \bibinfo{journal}{\emph{SCIENCE CHINA Information Sciences (SCIS)}}
  \bibinfo{volume}{1} (\bibinfo{year}{2024}), \bibinfo{pages}{1--14}.
\newblock


\end{thebibliography}

\clearpage
\newpage

\appendix
\section{Motivation}
	
	Camouflaged object detection (COD) often requires large-scale training and substantial computational resources. Although the Segment Anything Model (SAM) exhibits strong generalization, its performance on COD is limited by the high visual similarity between foreground and background. We observe that SAM benefits greatly from informative prompts, but manual prompt design is neither scalable nor efficient. To overcome this, we propose RAG-SEG, which leverages retrieval-augmented generation (RAG) to automatically extract prototypical features from the training set and use them as prompts for SAM. By fusing RAG’s ability to retrieve relevant patterns with SAM’s powerful segmentation backbone, RAG-SEG delivers accurate COD with minimal resource overhead. \textbf{Our objective in this paper is to demonstrate that RAG-SEG achieves high-quality segmentation using minimal computational resources, thereby highlighting the promise of RAG in the computer vision community.} In future work, we will incorporate supervised training to further advance performance toward state-of-the-art levels.

\section{ Experiment Settings}
All experiments were carried out on an affordable personal computer with the following specifications:
\begin{itemize}
  \item Processor: 11th Gen Intel(R) Core(TM) i5-11400H @ 2.70 GHz (2.69 GHz)
  \item RAM: 32.0 GB (31.8 GB available)
  \item Operating System: Windows 11, 64-bit architecture
  \item GPU: 4 GB NVIDIA GeForce 3050 Ti
\end{itemize}
The speed of RAG-SEG is remarkably fast.
Although our experiments could be conducted without relying entirely on the
CPU, we utilized an affordable GPU to accelerate the processes of feature extraction
and segmentation refinement in implementation. The experiments were conducted on a personal
device, where multiple applications, such as the Edge browser with over 200
open tabs, were running concurrently, utilizing both system RAM and GPU memory.
Therefore, the reported time may not be entirely accurate.
\textbf{We strictly followed the default settings of the COD benchmark without applying any modifications to the dataset or employing any data augmentation techniques.}

 \section{Availability of Some Qualitative Results}

 In the literature on training-free camouflaged object detection (COD), full visual examples are often difficult to obtain and quantitative results must be taken from the original publications. For example, two recent test-time adaptation frameworks—ProMaC and GenSAM—both rely on large foundation models and high-end GPUs, rendering them impractical on standard affordable laptop hardware.

\begin{itemize}
  \item \textbf{ProMaC} employs LLAVA-1.5-13B (GPT-4V), CLIP-CS ViT-B/16, and SAM, augmented by Stable Diffusion~2 for inpainting, and performs 4 iterations of \emph{iterative prompt refinement strategy} on a single NVIDIA A100 GPU.
  \item \textbf{GenSAM} integrates BLIP-2 (ViT-g OPT-6.7B), CLIP-CS ViT-B/16, and SAM, and conducts 6 iterations of \emph{iterative prompt refinement strategy} on a single NVIDIA A100 GPU.
\end{itemize}

 Both methods iteratively refine the segmentation by using the previous iteration’s mask as a prompt for the next. By contrast, our RAG-SEG requires only a single segmentation pass, achieves superior performance with minimal resources, and does not rely on test-time adaptation.  Moreover, neither ProMaC nor GenSAM currently release their qualitative segmentation outputs, and—owing to our hardware constraints—we were unable to reproduce their results. Upon acceptance, we will make our code and  visualizations publicly available on GitHub.
 
\section{Additional Comparison for COD}
\subsection{Additional Quantitative Comparison on NC4K}
As shown in Table~\ref{tab:nc4k_comparison}, our method requires the least computational resources yet outperforms MDSAM on the NC4K dataset, delivers a substantial improvement over DSAM, and achieves performance metrics comparable to those of COMPrompter. Note that NC4K results for some methods are not available and are therefore omitted.

\begin{table}[ht]
	\centering
	\caption{Quantitative comparison of various methods on the NC4K dataset. } 
	\label{tab:nc4k_comparison}
	\begin{tabular}{lcccc}
		\toprule
		\textbf{Method}              & $S_{\alpha}\uparrow$ & $E_{\xi}\uparrow$ & $F_{\beta}^{\omega}\uparrow$ & MAE$\downarrow$ \\
		\midrule
		SINet$_{2020}$               & 0.8080               & 0.7227             & 0.8713                       & 0.0576          \\
		SINetv2$_{2021}$             & 0.8472               & 0.7698             & 0.9027                       & 0.0476          \\
		COMPrompter$_{2024}$         & \textbf{0.9070}      &  0.8760    & \textbf{0.9550}              & \textbf{0.0300} \\
		MDSAM$_{2024}$               & 0.8750               &  \textbf{0.9210 }                & 0.8500                      & 0.0370          \\
		DSAM$_{2024}$                & 0.8710               & 0.8260             & 0.9320                       & 0.0400          \\
		RAG-SEG            & 0.8824               & 0.8507             & 0.9263                       & 0.0308          \\
		\bottomrule
	\end{tabular}
\end{table}

\subsection{Additional Visual Comparison Results}

\begin{figure*}[htpb]
  \centering
  \includegraphics[width=0.9\linewidth]{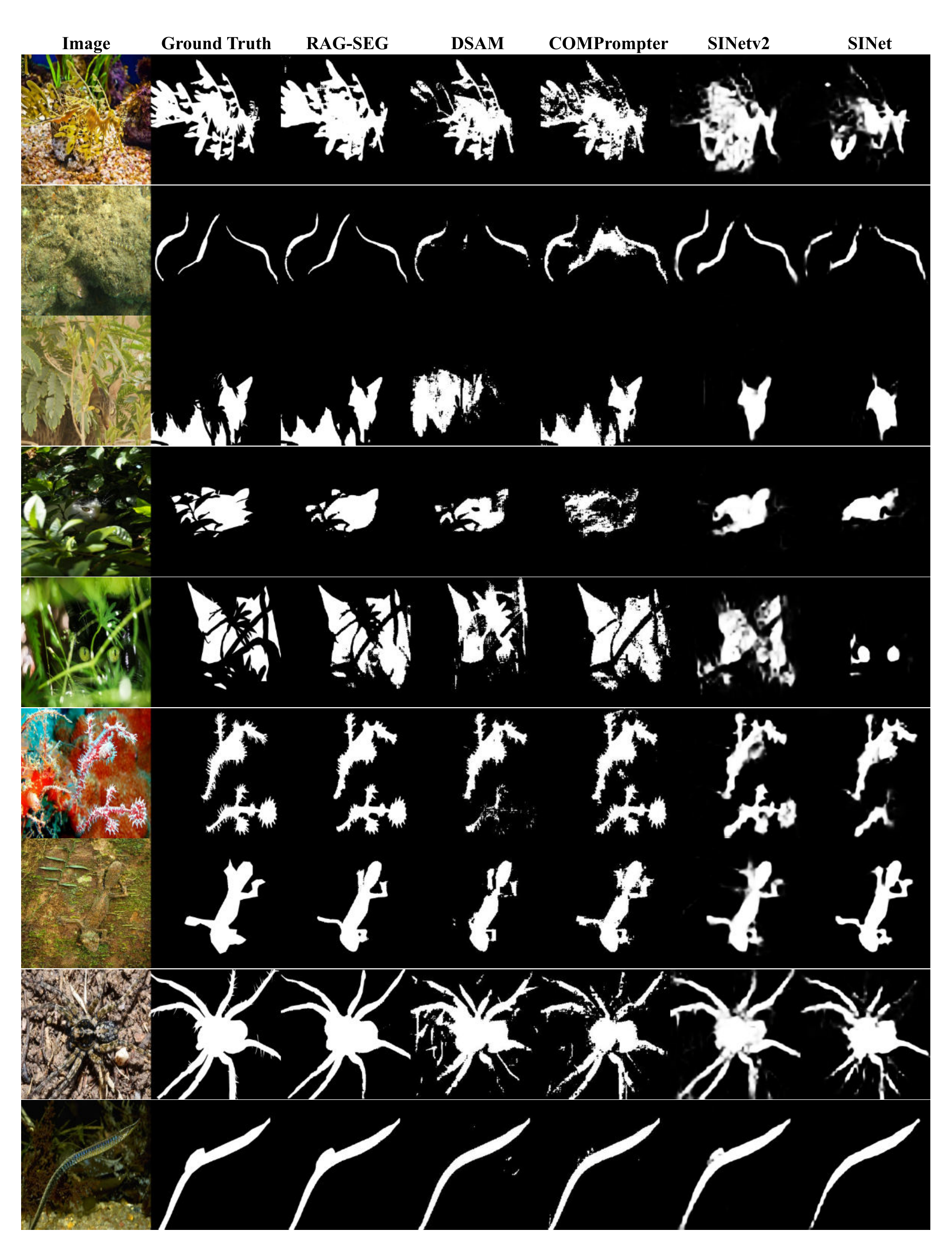}
  \caption{Additional visual comparison of the RAG-SEG framework against other SOTA methods.} 
  \label{fig:visual_comparison}
\end{figure*}

As shown in Figure~\ref{fig:visual_comparison}, our results demonstrate superior performance in scenarios involving occlusion, fine details, and multiple objects. This qualitative advantage helps explain why, despite not achieving state-of-the-art numerical scores compared to fine-tuned SAM methods, our approach produces more visually accurate and perceptually compelling segmentations.

\section{More Results about ablation studies}
  \subsubsection{Impact of Similarity Metric}
  In our experimental framework, we conducted evaluations based on a top-1
  criterion, wherein the system selects the most similar feature vector and its associated
  mask value. We examined three widely adopted similarity metrics: Inner Product
  (IP), Cosine Similarity, and L2 distance. The comparative performance of these
  metrics is shown in Table \ref{tab:similarity_comparison}. 
  The experimental results demonstrate that Inner Product and Cosine Similarity exhibit
  comparable performance characteristics, with IP showing marginally superior
  results across all evaluation metrics. 
  \begin{table}
  	\centering
  		\small
  	\caption{Performance comparison of different similarity metrics.}
  	\begin{tabular*}{\linewidth}{@{\extracolsep{\fill}}lcccc@{}}
  		\toprule 
  		\textbf{Metric} & $S_{\alpha}\uparrow$ & $F_{\beta }^{\omega }\uparrow$  &  MAE $\downarrow$  & $E_{\xi}\uparrow$     \\
  		\midrule 
  		Cosine          & 0.7534 & 0.6643 & 0.0866 & 0.7909 \\
  		IP              & \textbf{0.7570} & \textbf{0.6714} & \textbf{0.0856} & \textbf{0.7986} \\
  		L2              & 0.7468 & 0.6549 & 0.0901 & 0.7849 \\
  		\bottomrule
  	\end{tabular*}
  	
  	\label{tab:similarity_comparison}
  \end{table}
  \subsection{Impact of Feature Extractor}
\begin{figure*}[htpb]
	\centering
	\includegraphics[width=0.9\linewidth]{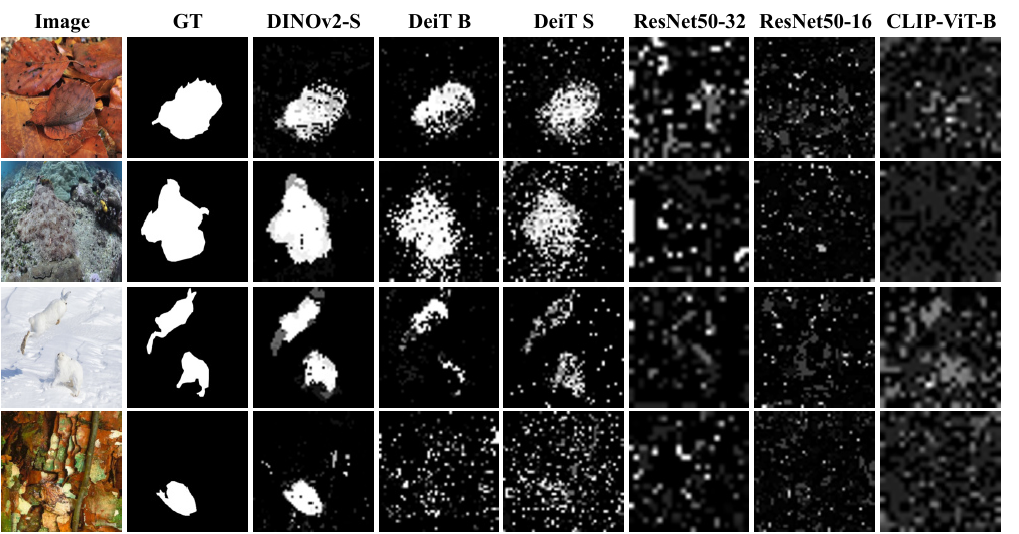}
    \caption{Additional visual comparison of RAG outputs produced by various feature extractors.}

	\label{tab:feature_extractor_visual}
\end{figure*}

   To validate our choice of DINOv2-S as the feature extractor in RAG-SEG, we conduct an ablation study under identical settings ($K=4096$, top-$k=1$), comparing four backbones: DeiT, CLIP-ViT, ResNet-50, and DINOv2-S. The quantitative results are presented in Table~\ref{tab:feature_extractors} while the visual results are in Figure ~\ref{tab:feature_extractor_visual}. Among these, DINOv2-S achieves the highest segmentation accuracy and the most precise localization of camouflaged targets, significantly outperforming DeiT, CLIP-ViT, and ResNet-50. This demonstrates that DINOv2-S provides the most discriminative and robust features for our RAG-SEG framework.
 \begin{table}
 	 	\centering
 			\small
 			\caption{Performance comparison of different Feature Extractors (FEs) of RAG-SEG ($K=4096$, topk = 1), where "B" and "S" refer to the Base and Small versions, respectively, and ResNet50-32/16 correspond to the features from the final and penultimate stages.}
 			\begin{tabular}{lcccc} 
 					\toprule 
 					\textbf{FE}& $S_{\alpha}\uparrow$ & $F_{\beta}^{\omega}\uparrow$ & MAE$\downarrow$ & $E_{\xi}\uparrow$ \\
 					\midrule 
 					CLIP-VIT-B & 0.4601 & 0.1226 & 0.1746 & 0.3567 \\
 					DeiT-B     & 0.6402 & 0.5525 & 0.1981 & 0.7011 \\
 					DeiT-S     & 0.5904 & 0.5171 & 0.2621 & 0.6496 \\
 					DINOv2-S   & \textbf{0.8334} & \textbf{0.8004} & \textbf{0.0579} & \textbf{0.8866} \\
 					ResNet50-16& 0.4199 & 0.0640 & 0.2057 & 0.3178 \\
 					ResNet50-32& 0.3441 & 0.1843 & 0.3972 & 0.4516 \\
 					Swin-Base &0.6324&0.5423&0.1986& 0.6970 \\
 					\bottomrule
 				\end{tabular}
 		
 			\label{tab:feature_extractors}
 \end{table}
 
 \subsection{Impact of Clustering Method}
 
 We chose FAISS KMeans for its optimized memory footprint and speed on large-scale, high-dimensional vectors. In contrast, Spectral Clustering, Agglomerative Clustering, and KMedoids all ran \textbf{out of memory}-first on a standard laptop and then even on a workstation with an NVIDIA RTX 4090D (24\,GB VRAM), 16 vCPUs, and 80,GB RAM—due to their high eigen-decomposition or quadratic/cubic complexity. FAISS KMeans, however, completed clustering into 4\,096 clusters in just about 433.85 seconds, demonstrating its superior scalability and efficiency.
  \subsection{Impact of top-\(k\) Retrieval}
\begin{figure*}[htpb]
		\centering 
		\includegraphics[width=0.9\linewidth]{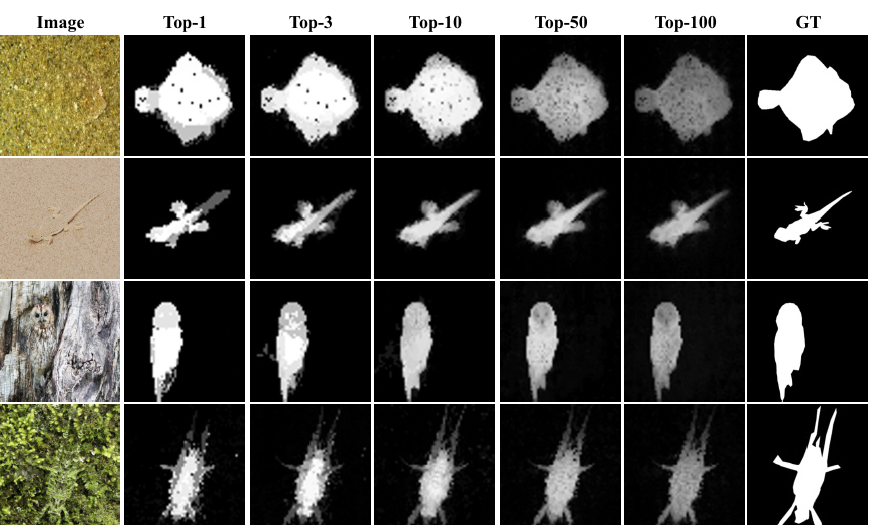}
		\caption{Visual comparison about initial mask across different top-\(k\) retrievals.} 
	
		\label{fig:top_k_visual}
\end{figure*}

  We evaluated the effect of varying Top-\(k\)  retrievals (\(k\)  = 1 to 10) on model performance,
  maintaining constant parameters of \(K\)=1024, Cosine metric, and $T_{3}$ strategy.
  Table \ref{tab:topk_comparison} presents our findings. The results demonstrate
  that lower k values consistently yield better performance across all metrics. Top-3
  retrieval achieved the highest $F_{\beta}^{w}$ (0.6692) and $E_{\xi}$
  (0.7938), while Top-1 performed best for $S_{\alpha}$ (0.7534) and
  $MAE$ (0.0866). Performance consistently declined for k values above 3.
  These results suggest that focusing on the most similar features (\(k\) $\leq$ 3) provides
  optimal mask prediction while maintaining efficiency. Figure~\ref{fig:top_k_visual} illustrates the initial masks generated by RAG for different top-\(k\) settings. As \(k\) increases, the masks exhibit reduced contrast and their pixel values converge toward 0.5, resulting in progressively blurrier segmentations. This degradation occurs because larger \(k\) injects more irrelevant features into the retrieval, amplifying noise and smoothing the mask values.
  
  \begin{table}[!htbp]
    \centering
    \small
    \caption{Performance comparison across different top-\(k\) retrievals.}
    \begin{tabular}{lcccc}
      \toprule $\mathbf{Top\text{-}k}$ & $S_{\alpha}\uparrow$ &  $F_{\beta }^{\omega }\uparrow$  & MAE$\downarrow$& $E_{\xi}\uparrow$     \\
      \midrule 1              & \textbf{0.7534}    & 0.6643                       & \textbf{0.0866}    & 0.7909                 \\
      3                       & 0.7541             & \textbf{0.6692}              & 0.0890             & \textbf{0.7938}        \\
      5                       & 0.7518             & 0.6652                       & 0.0903             & 0.7884                 \\
      10                      & 0.7444             & 0.6543                       & 0.0937             & 0.7817                 \\
      20                      & 0.7429             & 0.6521                       & 0.0951             & 0.7799                 \\
      50                      & 0.7368             & 0.6402                       & 0.0970             & 0.7726                 \\
      100 & 0.7270 & 0.6243 & 0.1011 & 0.7609 \\
      \bottomrule
    \end{tabular}
    
    \label{tab:topk_comparison}
   
  \end{table}
  \subsection{Analysis of Post-Processing Strategies}
  To enhance segmentation quality, we evaluated various post-processing
  strategies, with a specific focus on CRF. The experiments examined CRF
  application at two key stages: after retrieval-based prediction (Initial-CRF, I-CRF)
  and after SAM-based refinement (Final-CRF, F-CRF). The results are summarized
  in Table~\ref{tab:post_processing}. The ``CRF-ONLY" strategy applies CRF
  directly to the retrieval-based prediction for the final output, while the ``Baseline"
  strategy relies solely on SAM2 to generate predictions without additional
  processing.
  \begin{table}[htbp]
    \centering
    \small
     \caption{Comparison of CRF strategies on the CAMO dataset. }
    \begin{tabular}{lcccc}
      \toprule $\mathbf{Strategy}$ & $S_{\alpha}\uparrow$ &  $F_{\beta }^{\omega }\uparrow$  & MAE$\downarrow$& $E_{\xi}\uparrow$     \\
      \midrule CRF-ONLY          & 0.5366             & 0.3482                       & 0.1383             & 0.4995                 \\
      \midrule Baseline          & 0.7534             & 0.6643                       & 0.0866             & 0.7909                 \\
      +I-CRF                     & 0.6526             & 0.4909                       & 0.1178             & 0.6757                 \\
      +F-CRF                     & \textbf{0.7536}    & \textbf{0.6642}              & \textbf{0.0865}    & \textbf{0.7902}        \\
      +I-CRF + F-CRF             & 0.6529             & 0.4904                       & 0.1177             & 0.6730                 \\
      \bottomrule
    \end{tabular}
    \label{tab:post_processing}
    
  \end{table}
  The results reveal that CRF post-processing provides limited improvements in segmentation
  accuracy for camouflaged object detection. Applying CRF independently (\textit{CRF-ONLY})
  results in significantly lower performance ($S_{m}= 0.5366$), while combined
  strategies, such as Initial-CRF + Final-CRF, fail to outperform the baseline. These
  observations highlight the challenges of camouflaged object detection, where
  object boundaries often lack distinct intensity or texture features. The Final-CRF
  strategy achieves results comparable to the baseline, indicating that the
  benefit of CRF post-processing in this context is minimal. Consequently, CRF
  is excluded from our pipeline.

  \subsection{Comparison with Other Segmentation Models}
  To evaluate the segmentation component of our RAG-SEG framework, we conducted experiments on COD using MobileSAM, SAM, and SAM2 under the same condition.
  Despite
  achieving favorable results on some images, MobileSAM and SAM showed
  suboptimal performance on camouflaged objects when compared to SAM2.  This difference in performance can be attributed to SAM2’s stronger ability to generalize across different camouflage patterns and backgrounds.  This performance
  gap underscores the superiority of SAM2, which we chose as the secondary
  segmentation model in our study. The results of these models are summarized in
  Table~\ref{tab:segmentation_models_comparison}, where SAM2 significantly outperforms
  both MobileSAM and SAM. All models were evaluated using the same post-processing
  strategy, specifically the $T_{3}$ thresholding technique, and no additional techniques
  were applied. The input resolution for all models was set to 1024. 
  \begin{table}[htbp]
    \centering
    \small
     \caption{Segmentation performance of different models. }
    \begin{tabular}{lcccc}
      \toprule $\mathbf{Model}$ & $S_{\alpha}\uparrow$ &  $F_{\beta }^{\omega }\uparrow$  & MAE$\downarrow$& $E_{\xi}\uparrow$     \\
      \midrule MobileSAM      & 0.4072             & 0.0098                       & 0.1854             & 0.2696                 \\
      SAM               & 0.4085             & 0.0038                       & 0.1821             & 0.2570                 \\
      SAM2                    & \textbf{0.7534}    & \textbf{0.6643}              & \textbf{0.0866}    & \textbf{0.7909}        \\
      \bottomrule
    \end{tabular}
   
    \label{tab:segmentation_models_comparison}
  \end{table}

  \subsection{Impact of Positive and Negative Point Prompts}
  We conducted an ablation study to analyze the effect of varying positive and
  negative point thresholds and mask prompts on model performance. The results are
  presented in Table \ref{tab:prompt_comparison}, where we evaluate the
  segmentation performance using S-measure ($S_{m}$). The experimental results demonstrate
  that integrating both positive and negative point prompts, with appropriately
  chosen thresholds, significantly improves performance compared to using mask prompts
  alone.

  When using only mask prompts (first row in the table), the model achieves an
  $S_{m}$ score of 0.7534. In comparison, incorporating point prompts with a
  positive threshold of $T_{p}= 0.95$ and a negative threshold of $T_{n}= 0.005$
  yields a substantial improvement with an $S_{m}$ score of 0.8263. Similar performance
  is achieved with $T_{p}= 0.99$ and $T_{n}= 0.05$, resulting in an $S_{m}$
  score of 0.8254.

  The importance of combining both mask prompts and point prompts is further
  highlighted by the significant performance drop ($S_{m}= 0.7300$) observed
  when the mask prompt is removed while maintaining point prompts ($T_{p}= 0.95$,
  $T_{n}= 0.005$). Throughout our experiments, we maintained a consistent mask prompt
  threshold of 0.3, corresponding to the $T_{3}$ strategy. These findings demonstrate
  that a well-balanced integration of both point prompts and mask prompts is crucial
  for optimal segmentation performance.
  \begin{table}[htpb]
    \centering
    \small
     \caption{Ablation study on positive and negative point thresholds and mask
    	prompts. The mask prompt threshold is set to 0.3 (corresponding to the
    	$T_{3}$ strategy). Best results are highlighted in bold.}
    \begin{tabular}{lllc}
      \toprule $\mathbf{Positive\ T_{\text{p}}}$ & $\mathbf{Negative\ T_{\text{n}}}$ & $\mathbf{Mask}$ & $S_{\alpha}\uparrow$ \\
      \midrule -                         & -                         & 0.3           & 0.7534                      \\
      0.75                               & 0.005                     & 0.3           & 0.8251                      \\
      0.85                               & 0.005                     & 0.3           & 0.8174                      \\
      0.95                               & 0.005                     & 0.3           & 0.8263                      \\
      0.95                               & 0.005                     & -             & 0.7300                      \\
      0.95                               & 0.01                      & 0.3           & 0.8216                      \\
      0.95                               & 0.05                      & 0.3           & 0.8215                      \\
      0.95                               & 0.1                       & 0.3           & 0.8235                      \\
      0.99                               & 0.005                     & 0.3           & 0.8240                      \\
      0.99                               & 0.01                      & 0.3           & 0.8228                      \\
      0.99                               & 0.05                      & 0.3           & \textbf{0.8254}             \\
      0.99                               & 0.1                       & 0.3           & 0.8202                      \\
      \bottomrule
    \end{tabular}
  
    \label{tab:prompt_comparison}
  \end{table}

  \section{Scalability to Large-Scale Datasets}

Although standard camouflaged object detection (COD) benchmarks contain only 4,040 images, related tasks—such as salient object detection (SOD)—offer large-scale datasets with over 10,000 samples. To evaluate the scalability of RAG-SEG, we apply our framework to the large-scale DUTS-TR dataset, which contains 10,533 training images, and evaluate performance on the standard SOD benchmark DUTS-TE (5,019 images). The settings follow those used in the main paper.
\begin{table}[ht]
	\centering
	\small
	\caption{Impact of centroid count \(K\) on clustering time, per-image 784 \(\times\) 784 search latency, and storage footprint.}
	\label{tab:sod_clustering_comparison}
	\begin{tabular}{lccc}
		\toprule
		K & Clustering Time (s) & Search Time per Image (s) & File Size (MB) \\
		\midrule
		1024    &   37.4264   &   0.032836   &    1.62  \\
		4096    &  289.9503   &   0.061045   &    6.49  \\
		8192    &  925.0247   &   0.092390   &   12.97  \\
		10240   & 1422.8707   &   0.107845   &   16.21  \\
		16384   & 2333.6137   &   0.176441   &   25.94  \\
		32768   & 4682.4752   &   0.284777   &   51.88  \\
		65536   & 9377.0623   &   0.575370   &  103.75  \\
		\bottomrule
	\end{tabular}
\end{table}

Table~\ref{tab:sod_clustering_comparison} reports clustering time, average per-image search time (for 784\(\times\)784 resolution), and the corresponding file size for storing vector–mask pairs under different numbers of clusters \(K\). As seen, both clustering and search time scale approximately linearly with \(K\), reflecting the trade-off between efficiency and granularity of representation. Moreover, the modest file size underscores the efficiency of our approach and suggests its suitability for deployment on resource-constrained platforms such as mobile devices.
\begin{table}[ht]
	\centering
	\small
	\caption{Quantitative comparison on the DUTS-TE dataset for varying centroid counts $K$. } 
	\label{tab:duts_te_k_performance}
	\begin{tabular}{rcccc}
		\toprule
		\textbf{K} & $S_{\alpha}\uparrow$ & $F_{\beta }^{\omega }\uparrow$  & MAE$\downarrow$& $E_{\xi}\uparrow$     \\
		\midrule
		1024   & 0.8828 & 0.8478 & 0.0370 & 0.9213 \\
		4096   & 0.8863 & 0.8526 & 0.0354 & 0.9248 \\
		8192   & 0.8865 & 0.8520 & 0.0353 & 0.9224 \\
		10240  & 0.8906 & 0.8592 & 0.0331 & 0.9275 \\
		16384  & 0.8893 & 0.8571 & 0.0338 & 0.9269 \\
		32768  & 0.8936 & 0.8639 & 0.0328 & 0.9311 \\
		65536  & \textbf{0.8974} & \textbf{0.8701} & \textbf{0.0315} & \textbf{0.9331} \\
		\bottomrule
	\end{tabular}
\end{table}

\begin{figure*}[htpb]
  \centering
  \includegraphics[width=0.9\linewidth]{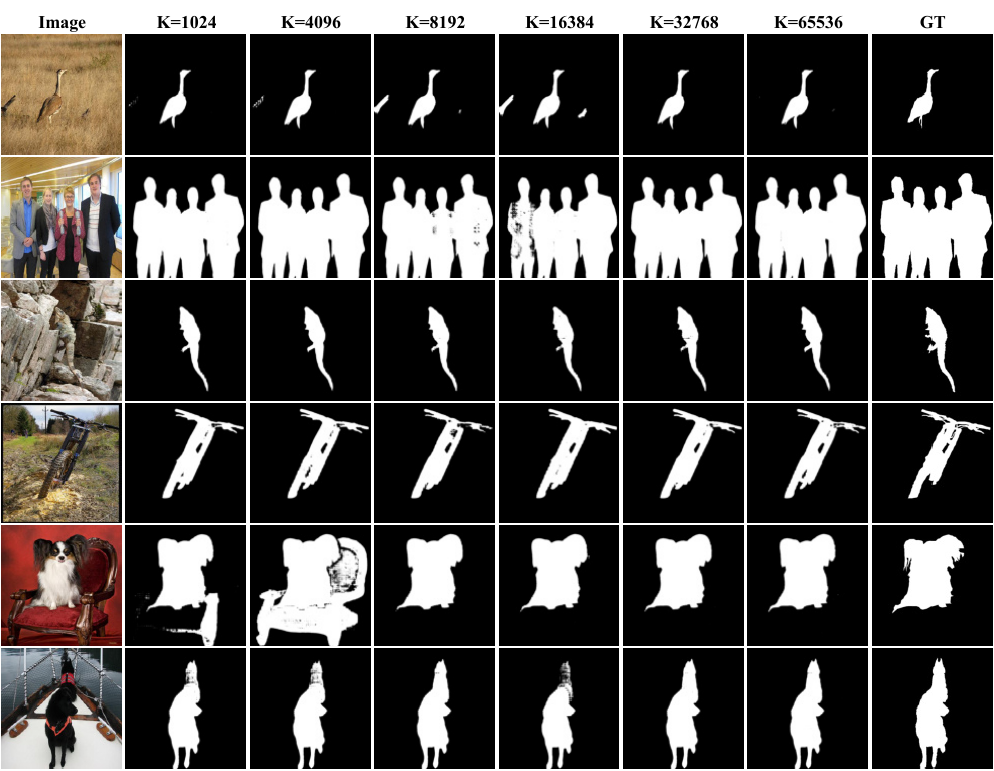}
  \caption{Visual comparison of the RAG-SEG framework for SOD with various \(K\).} 

  \label{fig:visual_cmp_sod_k}
\end{figure*}

Table~\ref{tab:duts_te_k_performance} presents the performance on DUTS-TE across different cluster sizes. Interestingly, increasing \(K\) does not consistently lead to improved results, although \(K=65536\) yields the best overall performance across metrics. Figure~\ref{fig:visual_cmp_sod_k} further supports this observation, showing that larger values of \(K\) do not significantly enhance segmentation quality. This phenomenon likely stems from the relative simplicity of the SOD task compared to COD: SAM's learned priors already enable effective localization of salient objects, but for SOD, SAM still requires appropriate prompts to suppress noisy or irrelevant masks. As a result, increasing \(K\) beyond a certain threshold offers diminishing returns.

In the SOD field, training-free approaches are basically rare. We compare RAG-SEG with MDSAM, a SAM-based method fine-tuned for SOD. \textbf{MDSAM is trained for 80 epochs on DUTS-TR using a SAM model fine-tuned on an NVIDIA A100 GPU.} Additionally, to further validate the effectiveness of RAG-SEG, we include a comparison with SAM2’s built-in \texttt{AutomaticMaskGenerator} (AMG) functionality. Since AMG produces multiple masks per image, we employ \textbf{ground-truth} filtering to retain only the target object masks. 

\textbf{As indicated in the official SAM2 repository, using multiple prompt types in AMG significantly increases memory consumption and inference latency.} Therefore, we test different prompt counts by uniformly sampling points along the image borders: 5, 8, 10 and 16 points per side, resulting in total prompts \(P=25\), \(64\), \(100\), and \(256\) respectively. These variants are denoted as \textbf{SAM2-AMG-P-GT}, where \(P\) indicates the total number of prompts.
\begin{table}[htbp]
	\centering
	\caption{Comparison of different methods on DUTS-TE dataset.}
	\label{tab:duts-te-results_cmp}
	\begin{tabular}{lcccc}
		\toprule
		\textbf{Method}& $S_{\alpha}\uparrow$ & $F_{\beta }^{\omega }\uparrow$  & MAE$\downarrow$ & $E_{\xi}\uparrow$     \\
		\midrule
		MDSAM\(_{2024}\)  (Trained)  & \textbf{0.9198} & \textbf{0.8928}      & \textbf{ 0.0245} & \textbf{0.9494} \\
		\midrule
		SAM2-AMG-25-GT  & 0.7250 & 0.6442 & 0.1179 & 0.7634 \\
		SAM2-AMG-64-GT  & 0.7827 & 0.7290 & 0.0813 & 0.8181 \\
		SAM2-AMG-100-GT & 0.8036 & 0.7605 & 0.0683 & 0.8375 \\
		SAM2-AMG-256-GT & 0.8247 & 0.7930 & 0.0564 & 0.8589 \\
		RAG-SEG (\(K\)=65536)         & 0.8974 & 0.8701 & 0.0315 & 0.9331 \\
		
		\bottomrule
	\end{tabular}
\end{table}
\begin{figure*}[htpb]
  \centering
  \includegraphics[width=0.8\linewidth]{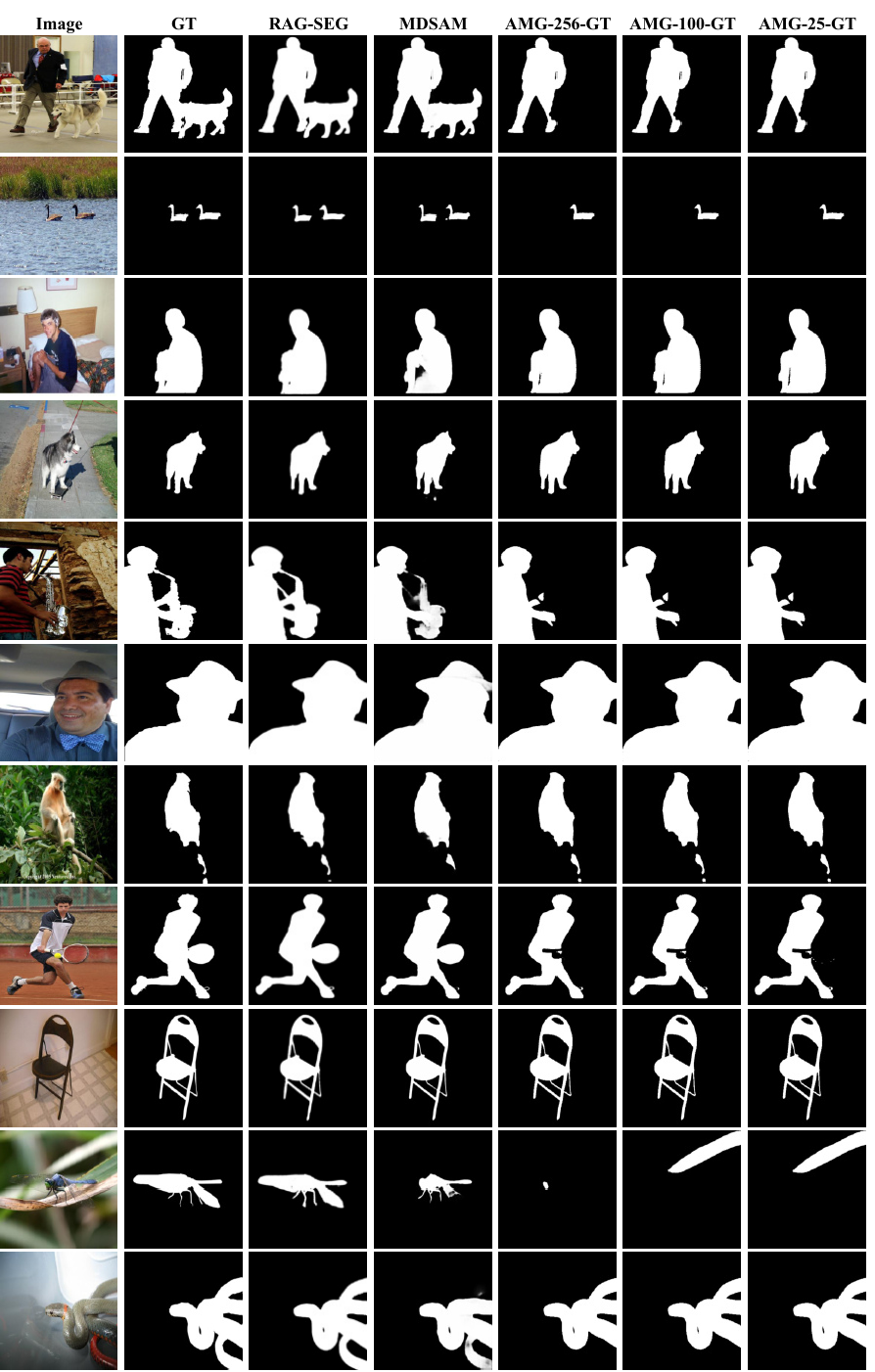}
  \caption{Qualitative comparison of RAG-SEG with existing SOD methods. We further include AMG-P-GT results, obtained by applying SAM2's AMG module with P point prompts (P = 25, 100, 256).} 
  \label{fig:vis_cmp_dutes}
\end{figure*}

Table~\ref{tab:duts-te-results_cmp} demonstrates that RAG-SEG substantially outperforms all AMG-based methods. Notably, our approach utilizing only 20 point and mask prompts surpasses the performance of AMG with GT filtering even when employing 256 points. \textbf{These results clearly illustrate SAM2's dependency on prompt quality and quantity. } When compared to MDSAM, which requires extensive computational resources for training specifically on SOD tasks, our training-free RAG-SEG approach achieves remarkably comparable performance across multiple evaluation metrics. As shown in Figure~\ref{fig:vis_cmp_dutes}, our method achieves better visual results than both the AMG-based methods and the fine-tuned MDSAM, demonstrating its effectiveness without any model adaptation.

In contrast to conventional SAM-based SOD pipelines—which typically require hundreds of ground-truth points to retrieve the best-matching mask and consequently suffer from substantial memory and computational costs—RAG-SEG generates compact prompt masks directly through retrieval, without any ground-truth supervision. As shown in Table~\ref{tab:duts-te-results_cmp}, RAG-SEG not only maintains high segmentation quality but also significantly reduces both memory footprint and inference time on large-scale datasets exceeding 10,000 images. 

Based on the above analysis, we argue that using the AMG method is highly inefficient and relies on ground truth masks, which are impractical to obtain in real-world scenarios. Considering our hardware constraints and the poor performance of AMG, we exclude it from further experiments. Instead, we compare our method with MDSAM on additional SOD datasets including DUT-OMRON, ECSSD, HKU-IS, and PASCAL-S, as shown in Table~\ref{tab:sod_supp_comparison}. The results demonstrate that our approach RAG-SEG achieves competitive segmentation performance, comparable to MDSAM, despite requiring no large-scale training. 

\begin{table*}[t]
\centering
\caption{Quantitative comparison with MDSAM on four standard SOD datasets.}
\begin{tabular}{@{}>{\footnotesize}l@{\hspace{1em}}cccccccc@{}}
\toprule
\textbf{Method} & \multicolumn{4}{c}{\textbf{ECSSD}} & \multicolumn{4}{c}{\textbf{PASCAL-S}} \\
\cmidrule(lr){2-5} \cmidrule(lr){6-9} 
& $S_{\alpha}\uparrow$ & $F_{\beta}^{\omega}\uparrow$ & MAE$\downarrow$ & $E_{\xi}\uparrow$
& $S_{\alpha}\uparrow$ & $F_{\beta}^{\omega}\uparrow$ & MAE$\downarrow$ & $E_{\xi}\uparrow$
\\
\midrule
MDSAM$_{\text{2024}}$ 
& \textbf{0.9483} &\textbf{ 0.9463 }& \textbf{0.0215 }&\textbf{ 0.9671 }
& \textbf{0.8820 }& 0.8510 & 0.0518 & 0.9167  \\
RAG-SEG               
& 0.9267 & 0.9275 & 0.0283 & 0.9546 
& 0.8784 & \textbf{0.8586 }& \textbf{0.0447 }& \textbf{0.9269} \\
\bottomrule
\end{tabular}

\begin{tabular}{@{}>{\footnotesize}l@{\hspace{1em}}cccccccc@{}}
\toprule
\textbf{Method} & \multicolumn{4}{c}{\textbf{HKU-IS}} & \multicolumn{4}{c}{\textbf{DUT-OMRON}}  \\
\cmidrule(lr){2-5} \cmidrule(lr){6-9}
& $S_{\alpha}\uparrow$ & $F_{\beta}^{\omega}\uparrow$ & MAE$\downarrow$ & $E_{\xi}\uparrow$ & $S_{\alpha}\uparrow$ & $F_{\beta}^{\omega}\uparrow$ & MAE$\downarrow$ & $E_{\xi}\uparrow$ \\
\midrule
MDSAM$_{\text{2024}}$ 
& \textbf{0.9414} & \textbf{0.9348 }& \textbf{0.0193} & \textbf{0.9691}
& \textbf{0.8783} & \textbf{0.8235} &\textbf{ 0.0387} & \textbf{0.9099} \\
RAG-SEG              
& 0.9169 & 0.9175 & 0.0259  & 0.9573 
& 0.8043 & 0.7185 & 0.0654& 0.8343 \\
\bottomrule
\end{tabular}
\label{tab:sod_supp_comparison}
\end{table*}

\section{Visualization of Clustering Results}

Although t-SNE is a common method for visualizing clustering outcomes, it is impractical in our case due to the massive number of feature vectors involved. While dimensionality reduction methods like PCA can be applied, they still cannot fully represent the cluster structure in a compact form. Instead, we propose a quantitative method to indirectly verify clustering quality by analyzing the mask scores associated with the cluster centers.

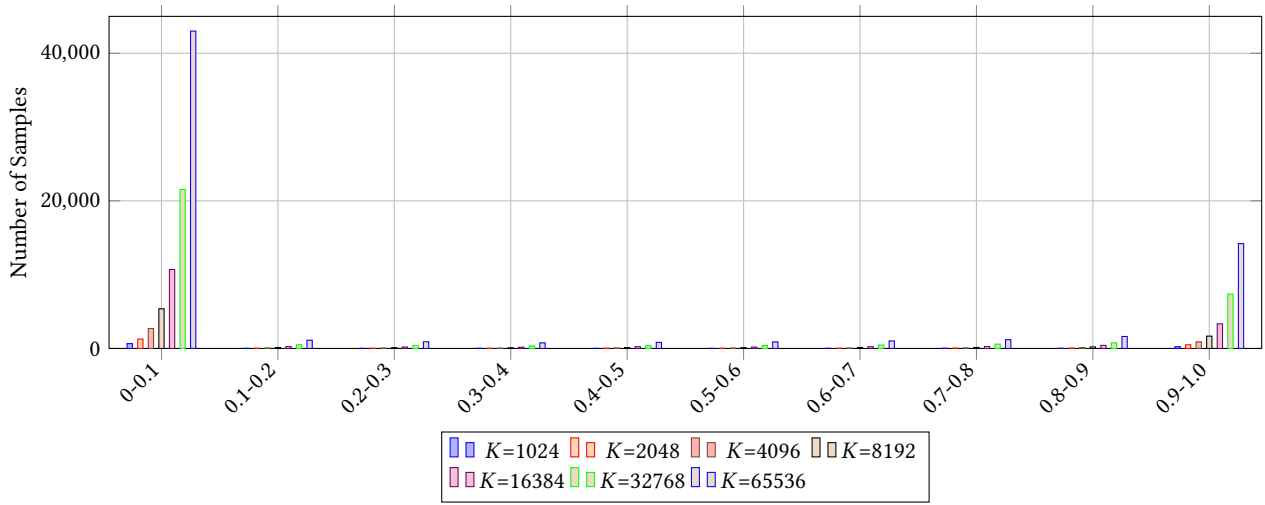
\begin{figure*}[htbp]
	\centering
	\begin{tikzpicture}
		\begin{axis}[
			ybar,
			bar width=2pt,
			width=0.95\textwidth,
			height=6cm,
			xlabel={Score Interval},
			ylabel={Number of Samples},
			symbolic x coords={
				0-0.1,0.1-0.2,0.2-0.3,0.3-0.4,0.4-0.5,
				0.5-0.6,0.6-0.7,0.7-0.8,0.8-0.9,0.9-1.0
			},
			xtick=data,
			x tick label style={rotate=45,anchor=east},
			legend style={at={(0.5,-0.25)}, anchor=north, legend columns=4},
			ymin=0,
			ymax=45000,
			scaled y ticks = false,
			yticklabel style={
				/pgf/number format/fixed,
				/pgf/number format/precision=0
			},
			ylabel style={yshift=-1mm},  
			enlarge x limits=0.05,
			grid=major
			]
			
			\addplot+[fill=blue!30] coordinates {
				(0-0.1,655) (0.1-0.2,16) (0.2-0.3,14) (0.3-0.4,14)
				(0.4-0.5,10) (0.5-0.6,10) (0.6-0.7,17) (0.7-0.8,23)
				(0.8-0.9,24) (0.9-1.0,241)
			};
			\addlegendentry{\(K\)=1024 }
			
			\addplot+[fill=orange!30] coordinates {
				(0-0.1,1273) (0.1-0.2,25) (0.2-0.3,23) (0.3-0.4,22)
				(0.4-0.5,24) (0.5-0.6,21) (0.6-0.7,27) (0.7-0.8,40)
				(0.8-0.9,46) (0.9-1.0,499)
			};
			\addlegendentry{\(K\)=2048}
			
			\addplot+[fill=red!30] coordinates {
				(0-0.1,2694) (0.1-0.2,63) (0.2-0.3,48) (0.3-0.4,39)
				(0.4-0.5,62) (0.5-0.6,46) (0.6-0.7,58) (0.7-0.8,72)
				(0.8-0.9,121) (0.9-1.0,893)
			};
			\addlegendentry{\(K\)=4096}
			
			\addplot+[fill=brown!30] coordinates {
				(0-0.1,5373) (0.1-0.2,126) (0.2-0.3,99) (0.3-0.4,84)
				(0.4-0.5,122) (0.5-0.6,95) (0.6-0.7,115) (0.7-0.8,134)
				(0.8-0.9,208) (0.9-1.0,1670)
			};
			\addlegendentry{\(K\)=8192 }
			
			\addplot+[fill=magenta!30] coordinates {
				(0-0.1,10700) (0.1-0.2,251) (0.2-0.3,196) (0.3-0.4,170)
				(0.4-0.5,241) (0.5-0.6,186) (0.6-0.7,235) (0.7-0.8,267)
				(0.8-0.9,418) (0.9-1.0,3339)
			};
			\addlegendentry{\(K\)=16384 }
			
			\addplot+[fill=brown!30] coordinates {
				(0-0.1,21544) (0.1-0.2,502) (0.2-0.3,403) (0.3-0.4,342)
				(0.4-0.5,390) (0.5-0.6,396) (0.6-0.7,470) (0.7-0.8,589)
				(0.8-0.9,770) (0.9-1.0,7362)
			};
			\addlegendentry{\(K\)=32768}
			
			\addplot+[fill=brown!30] coordinates {
				(0-0.1,43006) (0.1-0.2,1119) (0.2-0.3,906) (0.3-0.4,757)
				(0.4-0.5,824) (0.5-0.6,893) (0.6-0.7,1016) (0.7-0.8,1188)
				(0.8-0.9,1624) (0.9-1.0,14203)
			};
			\addlegendentry{\(K\)=65536}
			
		\end{axis}
	\end{tikzpicture}
	\caption{Distribution of SOD sample scores across 10 score intervals under different numbers of cluster centers..}
	\label{fig:score-distribution-multi_sod}
\end{figure*}

Each cluster center is assigned a mask score ranging from 0 to 1, reflecting its semantic saliency. To analyze the clustering quality, we partition this range into 10 uniform intervals and count the number of cluster centers within each bin. Figure~\ref{fig:score-distribution-multi_sod} presents the score distributions for different values of \(K\), highlighting how the number of high-score cluster centers increases with larger \(K\), indicating improved coverage of salient regions.

As shown, most cluster centers are concentrated in the lowest bin \([0.0, 0.1)\), particularly for large \(K\) values such as 32,768 or 65,536. This reflects that a majority of centers represent background or non-salient regions. However, while the number of high-score centers (e.g., \([0.9, 1.0)\)) does increase with larger \(K\), the improvement in clustering quality for foreground objects is relatively modest. This suggests that finer granularity provides some enhancement in the representation of salient areas, but the benefit to downstream segmentation tasks may not be as significant as expected.

\begin{figure*}[htbp]
	\centering
	\begin{tikzpicture}
		\begin{axis}[
			ybar,
			bar width=2pt,
			width=0.95\textwidth,
			height=6cm,
			xlabel={Score Interval},
			ylabel={Number of Samples},
			symbolic x coords={
				0-0.1,0.1-0.2,0.2-0.3,0.3-0.4,0.4-0.5,
				0.5-0.6,0.6-0.7,0.7-0.8,0.8-0.9,0.9-1.0
			},
			xtick=data,
			x tick label style={rotate=45,anchor=east},
			legend style={at={(0.5,-0.25)}, anchor=north, legend columns=3},
			ymin=0,
			ymax=7500,
			scaled y ticks = false,
			yticklabel style={
				/pgf/number format/fixed,
				/pgf/number format/precision=0
			},
			ylabel style={yshift=-1mm},
			enlarge x limits=0.05,
			grid=major
			]
			
			\addplot+[fill=blue!20] coordinates {
				(0-0.1,438) (0.1-0.2,3) (0.2-0.3,10) (0.3-0.4,2)
				(0.4-0.5,4) (0.5-0.6,7) (0.6-0.7,4) (0.7-0.8,4)
				(0.8-0.9,6) (0.9-1.0,34)
			};
			
			\addplot+[fill=blue!40] coordinates {
				(0-0.1,854) (0.1-0.2,9) (0.2-0.3,17) (0.3-0.4,13)
				(0.4-0.5,6) (0.5-0.6,9) (0.6-0.7,7) (0.7-0.8,11)
				(0.8-0.9,15) (0.9-1.0,83)
			};
			
			\addplot+[fill=blue!60] coordinates {
				(0-0.1,1727) (0.1-0.2,37) (0.2-0.3,27) (0.3-0.4,14)
				(0.4-0.5,14) (0.5-0.6,19) (0.6-0.7,18) (0.7-0.8,19)
				(0.8-0.9,20) (0.9-1.0,153)
			};
			
			\addplot+[fill=blue!80] coordinates {
				(0-0.1,3462) (0.1-0.2,54) (0.2-0.3,43) (0.3-0.4,44)
				(0.4-0.5,39) (0.5-0.6,35) (0.6-0.7,33) (0.7-0.8,40)
				(0.8-0.9,45) (0.9-1.0,301)
			};
			
			\addplot+[fill=blue!90] coordinates {
				(0-0.1,6879) (0.1-0.2,139) (0.2-0.3,104) (0.3-0.4,69)
				(0.4-0.5,61) (0.5-0.6,62) (0.6-0.7,78) (0.7-0.8,74)
				(0.8-0.9,115) (0.9-1.0,611)
			};
			
			\legend{512 centers, 1024 centers, 2048 centers, 4096 centers, 8192 centers}
		\end{axis}
	\end{tikzpicture}
	\caption{Distribution of COD sample scores across 10 score intervals under different numbers of cluster centers.}
	\label{fig:score-distribution-multi_cod}
\end{figure*}
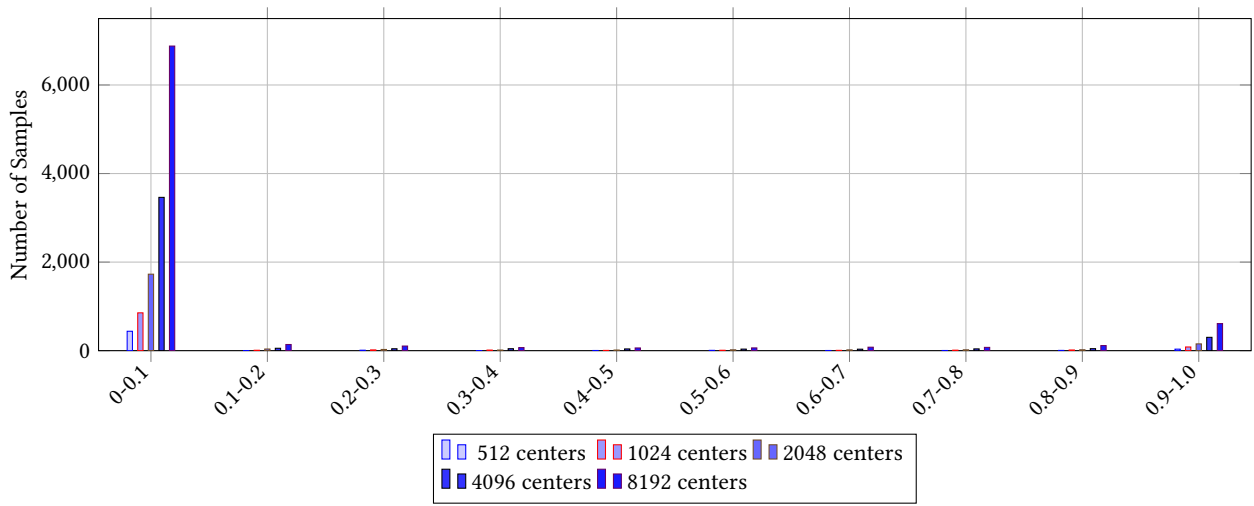

Figure~\ref{fig:score-distribution-multi_cod} presents the same score distribution as Figure~\ref{fig:score-distribution-multi_sod}. Notably, COD contains a much smaller proportion of cluster centers in the 0.9–1.0 score interval, indicating the scarcity of highly confident samples. This reflects the intrinsic difficulty of camouflaged object detection and also implies that camouflaged objects are generally smaller and less distinguishable than salient ones.

\section{Joint COD–SOD Segmentation}
The proposed RAG-SEG framework requires no task-specific training and can be applied directly to both camouflaged object detection (COD) and salient object detection (SOD). This raises the question of whether these two tasks can be addressed simultaneously within a single, training-free pipeline. Although most prior work treats COD and SOD as distinct problems, there is little exploration of unified, training-free solutions.
\begin{table*}[htpb]
  \centering
  \small
  \caption{Impact of joint clustering size \(K_U\) on COD and SOD performance on CAMO and ECSSD.} 
  \label{tab:joint_cod_sod}
  \begin{tabular}{lcccccc}
    \toprule
    Dataset & COD\;(K=4096) & SOD\;(K=65536)
      & $S_{\alpha}\uparrow$ & $E_{\xi}\uparrow$ 
      & $F_{\beta}^{\omega}\uparrow$ & MAE$\downarrow$ \\
    \midrule
     \multirow{2}{*}{CAMO}  
      & \checkmark & \multicolumn{1}{c}{}  
      & \textbf{0.8305} & \textbf{0.7950} 
      & 0.0637 & 0.8834 \\
      & \checkmark & \checkmark  
      & 0.8223 & 0.7886 
      & \textbf{0.0760} & \textbf{0.8802} \\
    \addlinespace[0.5em]  
    \multirow{2}{*}{ECSSD}  
      & \multicolumn{1}{c}{} & \checkmark  
      & \textbf{0.9267} & 0.9275 
      & \textbf{0.0283} & \textbf{0.9546} \\
      & \checkmark & \checkmark  
      & \textbf{0.9267} & \textbf{0.9280} 
      & 0.0281 & 0.9553 \\
    \bottomrule
  \end{tabular}
\end{table*}
To investigate this, we set \(K_{\text{COD}} = 4096\) and \(K_{\text{SOD}} = 65536\), and merge these into a combined dictionary size \(K_{U} = K_{\text{COD}} + K_{\text{SOD}}\). We then evaluate segmentation performance using each of \(K_{\text{COD}}\), \(K_{\text{SOD}}\), and \(K_{U}\). Notably, the unified dictionary shows negligible performance degradation and occasionally achieves slight improvements compared to the task-specific dictionaries (see Table~\ref{tab:joint_cod_sod}).
 We attribute this robustness to the intrinsic similarity of COD and SOD as binary segmentation tasks: COD can be viewed as a more challenging variant of SOD, and SOD as its simpler counterpart. Moreover, since our retrieval uses only the single most similar feature (top-1), the presence of features from the alternate task does not significantly disrupt matching. These findings demonstrate that \textbf{ RAG-SEG can serve as a truly unified COD–SOD segmentation method without any additional training} .
This outstanding property gives our method exceptional scalability: by iteratively generating segmentation masks on new images, we can extract features and progressively enhance the representations in our vector database. We will further explore and quantify this scalability in future work.
  	\section{Limitations}
	
	\subsection{RAG Process}
	\begin{itemize}
		\item \textbf{Fixed embeddings}: We rely on a pre-trained embedding model without fine-tuning, which may limit adaptability to new COD scenarios. Future work could explore fine-tuning or self-supervised refinement of the embedding space.
		\item \textbf{Basic RAG pipeline}: Our design employs a straightforward retrieval mechanism; advanced variants (e.g., GraphRAG) remain unexamined and could improve recall and precision.
		\item \textbf{Scalability}: Experiments were conducted on datasets of moderate scale. In Appendix~E, we further evaluate RAG-SEG on large-scale salient object detection benchmarks. For \textbf{ultra-large collections}, the adoption of inverted file (IVF) indexing or product quantization can preserve computational efficiency.
	\end{itemize}
	
	\subsection{SEG Process}
	\begin{itemize}
		\item \textbf{SAM dependency}: The framework performs two rounds of feature extraction with SAM and DINOv2, increasing computational load. Developing or fine-tuning a lightweight segmentation head could reduce redundancy and resource usage.
	\end{itemize}
  \section{Future Work}

This work introduces a novel exploration of training-free segmentation approaches
for COD. While our current implementation shows
promising results, we acknowledge certain limitations due to computational constraints
and the absence of model training in our approach. These limitations present
several promising directions for future research, which we outline below:

\textbf{Extension of the RAG-SEG framework to diverse segmentation tasks.}
Future research could expand the RAG-SEG paradigm beyond COD to encompass tasks such as semantic segmentation, panoptic segmentation,
and open-vocabulary segmentation. 

\textbf{Improvement of the RAG mechanism.} The current implementation relies on
a static retrieval system. Developing a dynamic RAG mechanism with an end-to-end
optimized retriever could significantly enhance retrieval accuracy and
efficiency, particularly for complex segmentation tasks.

\textbf{Revisiting vector storage architecture.} The list-like vector storage
structure used in this work is efficient but simple. Exploring graph-based
vector databases could better handle complex data relationships, improving
scalability and retrieval performance. 

\textbf{Optimization of segmentation models.} Leveraging lightweight network architectures or pre-trained models could enhance segmentation efficiency and reduce reliance on resource-intensive models such as SAM. This would accelerate the segmentation pipeline and enable deployment on edge devices with limited
computational capacity.

\end{document}